\documentclass[sn-basic,Numbered,iicol]{sn-jnl}

\usepackage{graphicx}%
\usepackage{multirow}%
\usepackage{amsmath,amssymb,amsfonts}%
\usepackage{amsthm}%
\usepackage{mathrsfs}%
\usepackage[title]{appendix}%
\usepackage{xcolor}%
\usepackage{textcomp}%
\usepackage{manyfoot}%
\usepackage{booktabs}%
\usepackage{algorithm}%
\usepackage{algorithmicx}%
\usepackage{algpseudocode}%
\usepackage{listings}%
\usepackage{xcolor}
\usepackage[normalem]{ulem} 

\makeatletter 
\def\@acknow{}%
\long\def\EarlyAcknow#1 \par{%
\def\@acknow{\abstractfont\abstracthead*{Acknowledgments}
#1\par}}%

\def\printabstract{\ifx\@acknow\empty\else\@acknow\fi\par%
    \ifx\@abstract\empty\else\@abstract\fi\par}
\makeatother

\newcommand\redsout{\bgroup\markoverwith{\textcolor{red}{\rule[0.5ex]{2pt}{0.4pt}}}\ULon}
\theoremstyle{thmstyleone}%
\theoremstyle{thmstyletwo}%

\theoremstyle{thmstylethree}%

\DeclareMathOperator{\comp}{c}
\DeclareMathOperator{\glob}{\#}
\DeclareMathOperator{\gr}{\mathcal{G}}
\DeclareMathOperator{\degr}{\tau}
\DeclareMathOperator{\cost}{\mathcal{C}}
\DeclareMathOperator{\risk}{\rho}
\begin{document}


\title[Article Title]{Deep Multi-Objective Reinforcement Learning for Utility-Based Infrastructural Maintenance Optimization}


\author*[1]{\fnm{Jesse} \spfx{van} \sur{Remmerden}}\email{j.v.remmerden@tue.nl}

\author[2]{\fnm{Maurice} \sur{Kenter}}\email{m.kenter@amsterdam.nl}

\author[2,3]{\fnm{Diederik M.} \sur{Roijers}}\email{d.roijers@amsterdam.nl}

\author[4]{\fnm{Charalampos} \sur{Andriotis}}\email{C.Andriotis@tudelft.nl}

\author[1]{\fnm{Yingqian} \sur{Zhang}}\email{yqzhang@tue.nl}

\author[1]{\fnm{Zaharah} \sur{Bukhsh}}\email{z.bukhsh@tue.nl}

\affil*[1]{\orgdiv{Information Systems IE\&IS}, \orgname{Eindhoven University of Technology}, \orgaddress{\street{De Zaale}, \city{Eindhoven}, \postcode{5600 MB}, \country{Netherlands}}}

\affil[2]{\orgname{City of Amsterdam}, \orgaddress{\street{Amstel 1}, \city{Amsterdam}, \postcode{1011 PN}, \country{Netherlands}}}

\affil[3]{\orgdiv{AI Lab}, \orgname{Vrije Universiteit Brussel}, \orgaddress{\street{Pleinlaan 9}, \city{Brussel}, \postcode{1050}, \country{Belgium}}}

\affil[4]{\orgdiv{Faculty of Architecture and the Built Environment}, \orgname{Delft University of Technology}, \orgaddress{\street{Jaffalaan 9a}, \city{Delft}, \postcode{2600 AA}, \country{Netherlands}}}
\EarlyAcknow{This research is partially supported by funding from the Flemish Government under the "Onderzoeksprogramma Artifici\"{e}le Intelligentie (AI) Vlaanderen" program, and the Dutch Research Council (NWO) and the municipality of Amsterdam under the Urbiquay program of the STABILITY (Grant NWA.1431.20.004) and LiveQuay (Grant NWA.1431.20.002) projects. Personal thanks to L. Smalbil, who helped review the paper.
}


\abstract{
In this paper, we introduce Multi-Objective Deep Centralized Multi-Agent Actor-Critic (MO-DCMAC), a multi-objective reinforcement learning (MORL) method for infrastructural maintenance optimization, an area traditionally dominated by single-objective reinforcement learning (RL) approaches. Previous single-objective RL methods combine multiple objectives, such as probability of collapse and cost, into a singular reward signal through reward-shaping. In contrast, MO-DCMAC can optimize a policy for multiple objectives directly, even when the utility function is non-linear. We evaluated MO-DCMAC using two utility functions, which use probability of collapse and cost as input. The first utility function is the Threshold utility, in which MO-DCMAC should minimize cost so that the probability of collapse is never above the threshold. The second is based on the Failure Mode, Effects, and Criticality Analysis (FMECA) methodology used by asset managers to asses maintenance plans. We evaluated MO-DCMAC, with both utility functions, in multiple maintenance environments, including ones based on a case study of the historical quay walls of Amsterdam. The performance of MO-DCMAC was compared against multiple rule-based policies based on heuristics currently used for constructing maintenance plans. Our results demonstrate that MO-DCMAC outperforms traditional rule-based policies across various environments and utility functions. 
}

\maketitle

\section{Introduction}\label{sec:introduction}
For any nation, a robust and functional infrastructure system is required for the efficient transportation of commercial goods, individuals, and essential services, such as clean water and electricity. This is evidenced by the distinct correlation between a country's Gross Domestic Product (GDP) and the level of development of its infrastructure \cite{lavee2011effect}. Given this direct relationship, it is vital to have a comprehensive maintenance strategy that ensures all infrastructural components are maintained at or above the minimum required service level and whereby unnecessary maintenance is avoided. Such strategic planning is crucial not only for the sustained economic performance of a nation but also for safeguarding the well-being and quality of life of its populace. 

The typical maintenance strategy for large infrastructural assets is to define either a proactive or reactive maintenance policy. A reactive maintenance policy entails that maintenance is only executed if the asset is almost failing or has failed. The main benefit of this strategy is that maintenance is not done unnecessarily and is therefore cost-efficient; however, Sawnson \cite{reactive_proactive} describes this strategy as a fire-fighting strategy for maintenance planning because maintenance is only done if an asset is almost failing or already failed. These failures could be catastrophic with infrastructural assets, such as the collapse of the Grimburgwal quay wall in Amsterdam \cite{grimburgwal} or the deadly collapse of the Morandi bridge in Genoa \cite{genoa_bridge}.

A proactive maintenance policy would prevent these failures by performing maintenance so that the asset will never deteriorate to a level it can fail. This comes with the downside that proactive policies are less cost-efficient than reactive ones. How much less depends on how well the proactive policy is formulated. An example of a proactive policy is a time-based maintenance strategy, where maintenance is performed at a set interval. Time-based maintenance is easy to implement but is, in most cases, not optimal because maintenance is even performed when the asset is still in a healthy state and is, therefore, more costly \cite{time-based-condition}. Condition-based maintenance alleviates this by only doing maintenance if an asset is in a specific condition. However, this warrants sufficient asset monitoring through inspections or other methods. To improve this monitoring, predictive maintenance is applied \cite{ZONTA2020106889}, whereby the future state of an asset is predicted or when a failure will occur, minimizing the required maintenance and inspection needed, both costly operations and with scarce capacity. A prescriptive maintenance policy uses these predictions to construct an optimal maintenance plan by not only planning maintenance to prevent a single failure but also for a longer time span in which multiple failures can occur. 

Many different methods can be used for prescriptive maintenance; some examples are integer linear programming \cite{milp_pavement, HNAIEN2016556} or genetic algorithms \cite{AllahBukhsh2020, earthquake_ga, GUAN2022130103}. In recent years, reinforcement learning has been shown to be a promising research direction for prescriptive maintenance \cite{ANDRIOTIS2019106483, ANDRIOTIS_paper_2,groupDavid, hamilton_markov, hrl_hamida, SKORDILIS2020106600}. With Reinforcement Learning (RL), the problem of prescriptive maintenance is formulated under a sequential decision-making setting. This means that every decision made does not just impact the immediate future but also has long-term effects. For instance, neglecting maintenance could lead to catastrophic failure or require extensive maintenance in the future. RL learns to optimally make decisions through trial-and-error, whereby at each decision step, it receives a singular reward signal, which tells how well it is performing. The goal is to maximize the cumulative reward signal. Other reasons RL is promising for infrastructural maintenance planning is its' ability to scale to larger assets, plan for decades in the future, and plan under uncertainty \cite{ANDRIOTIS2019106483, ANDRIOTIS_paper_2}. 

This ability to plan under uncertainty is essential because a key insight behind prescriptive maintenance is to explicitly take uncertainty about the condition state of the asset into account while deciding the best course of action. This uncertainty stems from the fact that the structural condition of most infrastructural assets is not fully observable due to their physical locations (e.g., bridge piles being submerged in water) and/or incomplete inspection due to the complexity of the asset's components. Taking this uncertainty into account is essential in forming an effective maintenance strategy. 

In addition to considering the uncertainty, it is essential to take the maintenance goal into account. This goal is, in most cases, formulated into a set of objectives for which the maintenance plan should be optimized, such as monetary cost, safety, or availability. However, comparing these objectives is a challenging task. For example, we can clearly calculate the maintenance cost for either a road or sewer pipes; conversely, we can not quickly determine the economic benefits of performing the maintenance on those roads and sewer pipes, either due to the number of economic actors that utilize a road or that it has not a clear economic benefit in case of the sewer pipes. Furthermore, these maintenance activities aim to achieve several objectives, including ensuring serviceability for roads and sewer pipes and maintaining road availability. Therefore, optimizing infrastructural maintenance requires a multi-objective approach where these objectives are weighted against each other.

Existing RL methods for prescriptive maintenance can only optimize for a single objective, whereas most real-world maintenance optimization problems are multi-objective. Therefore, we introduce \emph{Multi-Objective Deep Centralized Multi-Agent Actor-Critic (MO-DCMAC)} for multi-objective prescriptive maintenance planning. MO-DCMAC is based on the DCMAC \cite{ANDRIOTIS2019106483} and MOCAC \cite{Reymond2023} algorithms for maintenance planning and multi-objective reinforcement learning, respectively. MO-DCMAC learns how to construct a maintenance plan in the same sequence as is currently done by asset managers, whereby the score is how well the maintenance plan performs according to the utility used. 

If we consider infrastructural maintenance optimization as a multi-objective problem, we also need to consider what kind of scenario our method will be used. Hayes et al. \cite{Hayes2022} describe multiple scenarios for when a multi-objective approach is required. In most of these scenarios described by Hayes et al., the weighting of the objectives is not determined; however, asset managers who currently plan the maintenance know how to weigh these objectives with existing methodologies. One of these methodologies is the \emph{Failure Mode, Effects, and Criticality Analysis (FMECA)}~\cite{borgovini1993failure}, which evaluates the utility (criticality score) of various asset failure modes based on a combination of costs, risks, and other factors such as environmental effects, all integrated in a non-linear manner. Therefore, according to Hayes et al. \cite{Hayes2022}, we should consider the \textit{known utility function scenario} because we know how to scalarize all the objectives into a singular value.  

One can assume that if we know how to weigh objectives against each other, we could use existing reinforcement learning methods for the maintenance optimization of larger assets \cite{ANDRIOTIS2019106483, ANDRIOTIS_paper_2, hrl_hamida} because we can, with the utility function, scalarize the objectives to a singular reward value. However, these existing methods are undesirable for multi-objective maintenance optimization, even with a known utility function. One reason these methods are undesirable is that they require scalarization at every training step, whereas the utility function weighs the objectives for the whole maintenance plan. Moreover, in this paper, we consider FMECA and other non-linear methodologies as our utility function, which could lead to our maintenance optimization problem being intractable \cite{10.5555/2591248.2591251} if we use standard reinforcement learning for our problem.

Moreover, we formulate maintenance optimization for infrastructural assets as a \emph{multi-objective partially observable Markov decision process model (MOPOMDP)} \cite{soh2011evolving}. This MOPOMDP is formulated not only to handle uncertainty about the state of the components but also to model different reward signals that can be taken into account with a non-linear utility function. Additionally, we introduce a novel method for incorporating a probability value as a reward or objective. This reward value will ensure that episodic return is always a valid probability, which can be discounted. Lastly, we test MO-DCMAC on multiple environments where maintenance is planned for an infrastructural asset. For these environments, we use the historical quay walls of Amsterdam as a real-world use case (Sections~\ref{sec:smaller_quay_wall} and \ref{sec:larger_quay_wall}).   

\section{Background}\label{sec:background}
In the background section, we discuss the key topics to understand MO-DCMAC. We begin with explaining standard reinforcement learning (RL) and deep reinforcement learning (DRL). In this part, we will also focus on actor-critic methods, seeing that MO-DCMAC is one of them. Thereafter, we will explain multi-objective reinforcement learning (MORL) and how it differs from standard RL. For example, we will state the essential background on why standard RL methods can not be used for our multi-objective maintenance optimization problem even though we have a known utility function.

\subsection{Reinforcement Learning}
In the realm of maintenance optimization, addressing the challenge of determining the most effective sequence of maintenance actions over time is essential. This problem is inherently a sequential decision-making dilemma, where the objective is to optimize the scheduling and execution of maintenance activities to maximize system reliability and minimize costs. To achieve this, we want an agent to learn an optimal maintenance policy. An agent learns this by interacting within an environment, whereby the environment describes the possible maintenance actions and their effect. To frame this issue in a structured manner, we can formulate the maintenance optimization problem as a Markov decision process (MDP). 

A MDP is tuple $\mathcal{M}_{\text{MDP}}=\left \langle S, A, T, \mathbf{R}, \gamma, H \right \rangle$, in which:
\begin{itemize}
    \item $s_t \in S$ is set of states.
    \item $a_t \in A$ are all the possible actions.
    \item $T: S \times A \times S \rightarrow \left [0,1 \right ]$ is the transition function that describes the probability of transitioning from state $s_t$ to state $s_{t+1}$ if action $a_{t}$ is taken at timestep $t$.
    \item $\mathbf{R}: S \times A \times S \rightarrow \mathbb{R}$ is the reward function. The reward function maps the state $s_t$ and the taken action $a_t$ to a scalar reward value $r_t$. This reward value $r_t$ tells how well the agent is performing.
    \item The discount factor $\gamma \in [0,1 ]$ determines the significance of future rewards in the learning process.
    \item $H$ is the horizon, which indicates how long an episode will be.
\end{itemize}

With an MDP, we want to have the agent learn policy $\pi$ that maximizes the discounted cumulative reward called return:
\begin{align} 
    \pi^{*} &= \text{arg } \underset{\pi}{\text{max}}\:  \mathbb{E} \left [ R_{t} \mid \pi, s_{t} = s\right ] \nonumber \\
    &= \text{arg } \underset{\pi}{\text{max}}\:  \mathbb{E} \left [ \sum_{k=t}^{H}\gamma^{k}r_{k} \mid \pi, s_{t} = s \right ]\label{eq:policy_maximizes_cum_rew}, 
\end{align}
with $\pi^{*}$ being the optimal policy. The expected return of policy from taking action $a_{t}$ at state $s_{t}$ is described through the state-action value function (Q-function):
\begin{equation} \label{eq:q_function}
    Q^{\pi}{\left(s, a \right)} = \mathbb{E}\left [ R_{t} \mid \pi, s_t=s, a_t=a\right ],
\end{equation}
with the optimal Q-function being $Q^{*}{\left(s,a \right)}=\underset{\pi}{\max}\:Q^{\pi}{\left(s,a\right)}$. Similarly, the expected return from a given state $s_t$ is formulated through the value function (V-function),
\begin{equation} \label{eq:value_function}
    V^{\pi}{\left(s\right)} = \mathbb{E}\left [ R_{t} \mid \pi, s_t=s\right ].
\end{equation}

The aforementioned equations can be solved using tabular approaches where state-action or state value is stored for each possible state and action. However, as the number of states or actions grows, a tabular approach, like Q-learning \cite{Watkins1992}, does not scale well. Deep reinforcement learning (DRL) provides a scaleable solution. Deep Q-learning (DQN) \cite{dqn} uses a neural network to learn the state-action value (Equation~\ref{eq:q_function}). DQN and similar methods that learn a state or state-action value are called value-based methods. Another approach is policy-based methods. These methods directly learn a policy $\pi{\left (a_t \mid s_t \right)}$, which defines a probability distribution over the possible actions (Equation~\ref{eq:policy_maximizes_cum_rew}). A major downside of policy-based methods is that they encounter a lot of variance during training and are thus unstable without any modifications.

One approach to reduce variance in policy-based methods is using actor-critic methods. An actor-critic method consists of two parts: the actor, which learns a policy $\pi$, and the critic, which learns to estimate the future expected returns through either a state-action (Equation~\ref{eq:q_function}) or only a state (Equation~\ref{eq:value_function}). Besides reducing the variance, it also enables the actor to be updated at every timestep and reduces the variance in training. Advantage Actor-Critic, introduced by Mnih et al. \cite{a3c_paper}, further reduces the variance by updating through the advantage function, $A{\left(s,a \right )}=Q{\left(s,a \right )}-V{\left(s \right )}$, which measures how much better or worse an action $a_t$ is compared to the other actions at state $s_{t}$. The advantage function can also be approximated by using only the state value-function, by $A{\left(s_t,a_t \right )} \approx r_{t} + \gamma V{\left(s_{t+1}\right)} - V{\left(s_{t}\right)}$, allowing the critic to only use the state value function $V$, instead of also the state-action value function. The policy is then updated through the following loss function,

\begin{align} 
L{\left ( \pi \right)} =& - \sum_{t=0}^{H}A{\left (s_{t},a_{t} \right)} \log{\left (\pi_{\theta}{\left (a_{t} \mid s_{t} \right)} \right)} \nonumber \\
=& - \sum_{t=0}^{H}\left ( r_{t} + \gamma V_\psi{\left(s_{t+1} \right)} - V_\psi{\left(s_{t} \right) } \right ) \nonumber \\
& \cdot \log{\left (\pi_{\theta}{\left (a_{t} \mid s_{t} \right)} \right)}\label{eq:actor_critic_pol_update},
\end{align}

whereby $\theta$ are the weights for $\pi$, the actor, and $\psi$ the weights for $V$, the critic.

\subsection{Multi-Objective Reinforcement Learning}
Standard RL methods can only optimize a policy for single-objective problems, like optimizing for cost. As such, these methods fall short of many real-world problems in which multiple objectives are in play, such as infrastructural maintenance optimization. For example, in some instances, we cannot convert an objective to cost, such as safety, or it is a complex task to capture the output of asset reliability assessment methodologies, such as FMECA, to a reward function that can combine these objectives. Therefore, we will utilize \emph{multi-objective reinforcement learning} to optimize a policy for infrastructural maintenance optimization, such that we can use these assessment methodologies in the learning process.

In multi-objective reinforcement learning (MORL), the environment and interactions are formulated as a multi-objective Markov decision process (MOMDP) \cite{Hayes2022}. This is the tuple  $\mathcal{M}_{\text{MOMDP}}=\left \langle S, A, T, \vec{\mathbf{R}}, \gamma, H \right \rangle$. The only noticeable difference between a standard MDP and a MOMDP is the reward function $\vec{\mathbf{R}}: S \times A \times S \rightarrow \mathbb{R}^d $, which returns a vector of $d$ objectives, instead of a singular scalar value. A MOMDP can be reduced to a standard MDP if $d=1$ because the reward function would only return a singular reward value.

In standard RL, we can easily determine if a policy $\pi$ is performing better than another policy ${\pi}'$ if
\begin{equation}\label{eq:policy_compare}
    \mathbb{E}{\left [R_t \mid \pi, s_{t}=s \right]} > \mathbb{E}{\left [R_t \mid{\pi}', s_{t}=s \right]}.
\end{equation}
In contrast, this is not trivial in MORL. If we have $d=2$ and the expected return for $\pi$ is $\left(0, 10\right)$ and for ${\pi}'$ it is $\left (10, 0 \right )$, we can not decide which policy is performing better or even if they are performing equally because one objective might be significantly more important than the other objective. This complexity in comparing objectives is evident in infrastructural maintenance optimization. For instance, is it worth improving the probability of collapse from 2\% to 1\% if the cost doubles? Asset managers use methodologies like FMECA to assess the best maintenance policy. Therefore, we will focus on the \emph{utility-based approach} as described by Hayes et al. \cite{Hayes2022}, so we can include methodologies such as FMECA in the training procedure.

A utility function $u: \mathbb{R}^d \rightarrow \mathbb{R}$ is a mapping of the episodic return $\vec{R}=\sum_{k=0}^{H}\gamma^{k}\vec{r}_k$ to a singular value. The utility function needs to be \emph{strictly monotonically increasing}, meaning that if one of the objectives increases, the utility return can never decrease. It is nearly impossible to find an optimal policy if the utility is not monotonically increasing. This is because, in utility-based MORL, the agent aims to learn a policy that maximizes the utility. However, if the utility function is not monotonically increasing, the agent will reach a certain point during training where it can only improve the utility further if it first decreases. This will cause the agent to get stuck at a local optimum.

There are two methods of optimizing over a known utility function in MORL \cite{Hayes2022}, namely the \emph{Scalarized Expected Return} (SER) criterion, which is when the utility function is applied on the expected return:
\begin{equation} \label{eq:ser_criterion}
    \pi^{*} = \text{arg } \underset{\pi}{\text{max}}\: u \left ( \mathbb{E} \left [ \sum_{k=t}^{H}\gamma^{k}\vec{r}_{k} \mid \pi, s_t=s\right ]\right).
\end{equation}
The other criterion is the \emph{Expected Scalarized Return} (ESR), which tries to maximize the expected return of the utility function:
\begin{equation} \label{eq:esr_criterion}
    \pi^{*} = \text{arg } \underset{\pi}{\text{max}}\:  \mathbb{E} \left [ u \left (\sum_{k=t}^{H}\gamma^{k}\vec{r}_{k} \right)\mid \pi, s_{t}=s \right ].
\end{equation}
The difference is that while SER is concerned with the utility of the average outcome, ESR takes the utility over every single roll-out of the policy (and only then takes the average). When we examine most MORL methods, we see that the SER criterion is more used, while in multi-objective games, the ESR criterion is used \cite{Hayes2022}. One explanation for the ESR criterion being understudied is that it invalidates the Bellman equation if the utility is non-linear:

\begin{align} 
    \underset{\pi}{\max}\: &\mathbb{E}{\left [u{\left (\vec{R}_{t}^{-} + \sum_{k=t}^{\infty}\gamma^{k}\vec{r}_k \right )} \mid \pi, s_{t} \right]} \neq \nonumber \\ &u{\left (\vec{R}_{t}^{-} \right )}+\underset{\pi}{\max}\: \mathbb{E}{\left [u{\left ( \sum_{k=t}^{\infty}\gamma^{k}\vec{r}_k  \mid \pi, s_{t}\right )} \right]},\label{eq:esr_bellman_invalid}
\end{align}
where $\vec{R}_{t}^{-}=\sum^{t-1}_{k=0}\gamma^{k} \vec{r}_{k}$ is the accrued reward, until timestep $t$. Equation~\ref{eq:esr_bellman_invalid} shows how the Bellman equation becomes invalidated because any non-linear utility function under the ESR criterion does not distribute across the accrued reward and the future returns \cite{Hayes2022}. Therefore, the most evident criterion to use would be the SER criterion; however, a policy learned with the SER criterion could differ significantly from one learned with the ESR criterion, as shown by R\u{a}dulescu et al. \cite{Radulescu2019}. Moreover, Hayes et al. \cite{Hayes2022} describe when ESR or SER should be used. SER is more flexible because it takes the expected return of the objectives as input for the utility function. This means that in our context of maintenance optimization, an asset might collapse in some runs so long it is safe on average. ESR is more strict because it optimizes the expected return of the utility function, which means that at every run, the asset should not collapse because, otherwise, the utility will be low. This strictness of the ESR criterion is more desired for maintenance optimization because an asset should never collapse.

\section{Related Work} \label{sec:related_work}
In infrastructural maintenance optimization, we can optimize either at the network level \cite{groupDavid, GUAN2022130103} or the asset level \cite{ANDRIOTIS2019106483,ANDRIOTIS_paper_2,cha_hrl, hrl_hamida}. At the network level, we optimize the maintenance for multiple assets, whereby each maintenance action is applied to the whole structure. When we plan maintenance at this level, we mainly consider the interaction between the assets. One of these interactions could be to ensure that destinations are reachable in a road network \cite{GUAN2022130103} or by rewarding the grouping of maintenance actions of assets close together \cite{groupDavid}. At the asset level, the level of our problem, we only consider a single asset but plan the maintenance of its components. Similar to the network level, we consider interactions, but instead between those components, such as a higher risk of failure when multiple critical components are failing.

When we compare other research, we notice that most asset level maintenance optimization problems are formulated as a POMDP. An example of this is DCMAC, by Andriotis and Papakonstantinou \cite{ANDRIOTIS2019106483}, which uses a discrete state space for the health level of each component, whereby the belief is calculated through a Bayesian update. It is also possible to use a continuous state space with a POMDP; however, this introduces increased complexity because the belief can not be updated anymore through a Bayesian update. The reason for this is that it requires an integral that has no closed-from solution. Nevertheless, it is still possible to use a continuous POMDP \cite{hamilton_markov, hrl_hamida, SKORDILIS2020106600}, in combination with approximation methods for the belief, such as Kalman Filters \cite{kalman}, Particle Filters \cite{particle_filter}, other Deep Learning methods \cite{drqn}, or other methods. However, in this paper, we will only consider a discrete state space and thus not require any of these methods.

Many methods are used for maintenance optimization besides deep Reinforcement Learning. Common methods are integer linear programming \cite{milp_pavement, HNAIEN2016556}, genetic algorithms \cite{AllahBukhsh2020, earthquake_ga, GUAN2022130103} and other optimization methods \cite{windmill_bee}. Nevertheless, these methods do not consider partial observability and struggle with scalability.

This struggle with scalability is also the reason why we focus on DRL methods instead of other methods, such as non-deep RL, that can handle partial observability, such as POMCP \cite{pomcp} or even methods for MOPOMDP \cite{roijers2015point}. In comparing the benchmarks used by both methods, we found that Roijers et al. \cite{roijers2015point} used benchmarks with significantly smaller state and action spaces. Some of the benchmarks used by Silver and Veness \cite{pomcp} had a larger state space than we used in our experiments; however, the action space is magnitudes smaller, with at most 100 possible actions, while our smallest environment has 13122 possible action combinations and the largest approximately $1.88 \times 10^{11}$. Therefore, the Monte-Carlo tree search used by Silver and Veness is infeasible with such a large action space.

This issue with the action space, or the "curse of dimensionality" of the action space, is a common problem when it comes to using RL for maintenance optimization problems because the agent should not only take a singular action for each timestep but rather an action for each component, such that the action space grows exponential to the number of components. In our research, we notice that two kinds of methods are used to address this issue. The most common method is to model it as a multi-agent problem, such as DCMAC \cite{ANDRIOTIS2019106483}. In it, Andriotis and Papakonstantinou use a factorized action space representation and model the neural network such that it has a shared input and a separate output head for each agent/component. The shared input is not always necessary, and DDMAC uses a separate input for each component \cite{ANDRIOTIS_paper_2, morato}. However, this has the limitation in that there is no communication between the agents/components, and each action will be taken independently of the states of other components. Another more recent approach is to use hierarchical reinforcement learning (HRL) \cite{cha_hrl, hrl_hamida} for tackling larger action spaces by decomposing them into smaller action spaces using HRL. However, we focus more on the factorized action space approaches in this paper.

In most MORL research, we see the focus on methods for training a Pareto Frontier of different policies \cite{envelope_q, Alegre2023SampleEfficientML}. However, these objectives are weighted linearly, through different weights, and use the SER criterion. Nevertheless, these methods are useful if the utility is unknown. These Pareto Frontier methods are already used for use-cases, such as the job-shop scheduling problem \cite{morl_jssp} and manufacturing \cite{manu_morl}.

As described by Hayes et al. \cite{Hayes2022}, the ESR criterion is understudied, which is mainly due to the increased complexity that comes with the ESR criterion and a known non-linear utility function. One method for this is multi-objective categorical actor-critic (MOCAC) \cite{Reymond2023}. During our research on applying ESR criterion deep MORL methods like MOCAC, we could not find any existing papers. Hence, to the best of our knowledge, this paper is the first time a deep MORL method with the ESR criterion has been employed to solve a real-world problem, such as optimizing infrastructural maintenance.

\section{Maintenance Optimization Problem} \label{sec:maintenance_problem}
Before we explain our proposed method, MO-DCMAC, we first state the problem formulation of the maintenance optimization problem, for which we want to find an optimal policy and how this is formulated as a \emph{Multi-Objective Partially Observable Markov decision process} (MOPOMDP) \cite{soh2011evolving}. In our explanation, we will also include how a MOPOMDP differs from a MOMDP and MDP.

\subsection{Problem Formulation}

An asset consists of $N$ components, whereby each component $\comp_i, i\in \{0,1,\dots,N-1\}$ can belong to one of $K \in \mathbb{N}$ component groups. The health state of each component $\comp_i$ is represented by an integer variable $s_t^i \in \{0,1,\dots, |S|-1\}$, where $|S|$ is the number of possible health states of the component. Each component $\comp_i$ has a current degradation rate $\degr_{t}^{i}$ that can be increased until $\degr_{\max}$. The degradation rate $\degr_{t}^{i}$ dictates the probability that component $\comp_i$ transitions from a healthy to a non-healthy state.  Besides the degradation rate, to what component group $\comp_i \in K$ belongs also influences the transition probability. This means some component groups might degrade faster than others, even if their degradation rate $\tau$ is higher.  

For each component $\comp_{i}$, we can take an individual maintenance action $a^{\comp_{i}}_t$, which is the action taken at timestep $t$. Each maintenance action $a^{\comp_{i}}_t$ will always increase the health state, except if the action is do nothing. If replaced, the degradation rate of the component $\comp_{i}$ is set back to 0. Besides taking an action for each component, we can also take an action $a^{\glob}_t$ for the whole asset. These \textit{global} actions will affect all the asset components and could be, for example, an inspection. Moreover, each action, for the component and global, will have a cost $c$, which is the monetary value of that action.

This maintenance optimization problem is \emph{partial observable}, so rather than observing the state $s^{\comp_{i}}_{t}$ of component $\comp_{i}$, we instead observe an observation $o_{t}^{\comp_{i}} \in \mathcal{O}$, where $\mathcal{O}$ is the set of possible observations. The observation $o_{t}^{\comp_{i}}$ is dependent on both the current health state $s^{\comp_{i}}_{t}$ of component $\comp_i$ as well as action taken at time $t$. For example, if a global inspection is done at timestep $t$, the next observation will fully reveal the state, meaning that $o_{t+1}=s_{t+1}$; however, if no inspection is done, but the component $\comp_i$ is repaired, then only the state is revealed for component $\comp_i$. If neither an inspection nor a maintenance action is done, the observation will be a low-accuracy representation of the current state. This accuracy depends on the specific maintenance environment.  

Within the $K$ components groups, multiple sub-dependency groups can exist $\gr^{k}_{i}$. Within the sub-dependency groups, there may be one or more components from the same component group. When one or more $\comp \in \gr^{k}_{i}$ are in the worst state $|S| - 1$, a probability $F{\left (\gr^{k}_{i} \right)}$ is returned that signifies the effect that $\gr^{k}_{i}$ has on \emph{probability of collapse}, which increases as more components are failing in the same component group. For example, if components $\comp_i \in \gr^{k}_{i}$ and $\comp_j \in \gr^{k}_{i}$ are both failing then $F{\left (\gr^{k}_{i} \right)}$ will be higher. However, if $\comp_j \notin \gr^{k}_{i}$ then $F{\left (\gr^{k}_{i} \right)}$ would not increase because component $\comp_j$ does not belong to the same component group.   

To calculate the probability of collapse $\risk_t$ for a single time step, we use the following equation that combines the effect of each component group on the probability of collapse:
\begin{equation} \label{eq:failure_prob_statement}
    \risk_t = 1 - \prod_{k\in K}\prod^{\left | \gr^{k} \right |-1}_{i=0}\left ( 1 - F{\left (\gr^{k}_{i} \right)} \right ).
\end{equation}

The objective is to find an optimal maintenance strategy, i.e., the series of maintenance actions until the horizon $H$, that minimizes both the total \emph{cost} $\cost$ and the \emph{probability of collapse} $\risk$ of the asset.
\subsection{Multi-Objective Partially Observable Markov Decision Process (MOPOMDP)}
In our problem formulation, we stated that the real health state of each component is only \textit{partially observable}. Any agent would likely not learn an optimal policy if we would formulate the problem with an MOMDP. Therefore, we must formulate the problem as \emph{Multi-Objective Partially Observable Markov decision process} (MOPOMDP) \cite{soh2011evolving}. A MOPOMDP is a tuple $\mathcal{M}_{\text{MOPOMDP}}=\left \langle S, A, \Omega, T, O, \vec{\mathbf{R}}, \gamma, H \right \rangle$, whereby a MOPOMDP adds the following:
\begin{itemize}
    \item $o_t \in \Omega$ is the set of observations that the agent can receive.
    \item $O: S \times A \times \Omega \rightarrow \left [0,1 \right]$ is the observation function and describes the probability that an observation is received based on the current state and action.
    
\end{itemize}
In a MOPOMDP, the agent receives an observation $o_t$ at each timestep $t$ instead of the current state $s_t$, whereby the observation is an inaccurate representation of that state. For example, sensors are inaccurate and could give different readings for the same state. Within a MOPOMDP, the sensor would be the observation function, and the sensor outputs the observation the agent would receive.

Due to the inaccuracy of the observation, a standard RL method will most likely result in a sub-optimal policy. A solution for this is to add some form of memory that keeps a belief over the current state $\mathbf{b}_t$. If the state space is discrete, the belief can be formulated as a vector of size $\left | S \right |$, whereby the belief is the probability that the agent is in that given state. This belief is then updated at each timestep through a Bayesian Update:

\begin{align}
        b{\left (s_{t+1} \right )} =& P \left (s_{t+1} \mid o_{t+1}, a_t, \mathbf{b}_t \right)\nonumber \\
        =& \frac{P \left (o_{t+1} \mid s_{t+1}, a_t\right )}{P \left ( o_{t+1} \mid \mathbf{b}_t, a_t \right )}\nonumber \\
        &\sum_{s_{t} \in S} P \left (s_{t+1} \mid s_t, a_t \right) b{\left ( s_t\right )}  \label{eq:belief_update}       
\end{align}

In the belief update, shown in Equation~\ref{eq:belief_update}, the belief $\mathbf{b}_t$ is updated through multiple mechanics. One of them is through the known transition function $P\left (s_{t+1} \mid s_{t}, a_{t} \right)$ whereby, even though if do not get any more information, add some more accurate information about the state. Secondly, the current action $a_{t}$ can also influence what kind of observation is received at the next timestep $t+1$, such as doing an inspection, which could give a more accurate observation.

The specific definition of our MOPOMDP is as follows:
\begin{itemize}
    \item $s_t \in S$ The state-space is discrete, with the $N$ components being able to be in one of $\left | S^{\comp_i} \right | = 5$ states and a degradation rate $\degr_{t}^{\comp_i}$ for each component. Whereby the maximum degradation rate is $\degr_{\max}=50$
    \item $a_t \in A$ The action-space is the combination of actions for the $N$ component, which is $A^{\comp_i} \in \left \{\textit{do nothing}, \textit{repair}, \textit{replace} \right \}$ with $1 \leq i \leq N$, and the global action for the whole structure, which is $A^{\glob} \in \left \{\textit{do nothing}, \textit{inspect} \right \}$. 
    \item $o_t \in \Omega$ The observation-space is equal to the state-space, except $\degr^{\comp_i}$, in that the observation for each component can be one of $\left | S^{\comp} \right |$ states.
    \item $T: S \times A \times S \rightarrow \left [0, 1 \right ]$ At each timestep $t$, the component has a probability of either staying in a specific state or transitioning to a worse state, based on $\degr^{\comp_i}_t$ and the associated transitioning matrix. If a \textit{repair} action is taken, the component condition is improved, which is depicted by moving one state up, and if the \textit{replace} action is taken, the component is set to the best state, and $\degr^{\comp_i}_{t+1}$ is set to 0.
    \item $O: S \times A \times \Omega \rightarrow \left [0, 1 \right ]$ The received observation is based on $a^{\glob}_{t}$. If the \textit{nothing} action is taken, the observation will be of low accuracy representation of the state. This low accuracy observation only indicates if a component is either in the first two states or the last three states. Still, it cannot differentiate between those specific states. If the \textit{inspect} action is taken, the state will be fully observed. Moreover, the state of a specific component $\comp_{i}$ is also fully observed if $\comp_{i}$ if a \textit{repair} or \textit{replace} action is taken for $\comp_{i}$.
    \item $\vec{R}: S \times A \times S \rightarrow \mathbb{R}^{2}$ The reward vector consists of two values $\vec{r}_t = \{-\cost_t, \log{\left ( 1 - \risk_{t}\right )}\}$. $-\cost_t$ represents the \textit{cost}, which has been made negative because a lower cost is preferable. $\log{\left ( 1 - \risk_{t}\right )}$ is the reward value for the \textit{probability of collapse}. This reward value represents the inverse log probability of collapse occurring at timestep $t$. Using this log probability as our reward value, we can calculate the probability of collapse of the whole episode as input for our utility function. We explain this in-depth later in this section.
    \item $H$ The horizon of any maintenance optimization is the time window and interval when maintenance should be planned. For example, if maintenance is planned yearly for 50 years, $H=50$. If the interval is monthly, then the horizon for 50 years is $H=600$. However, in this paper, $H=50$ is used because we only consider yearly intervals for a time window of 50 years. 
\end{itemize}
Due to the partial observability of the state, we introduce a belief $\mathbf{b}_{t}$ over the states. We update this belief with the newly received observation $o_{t}$ through a Bayesian update as seen in Equation~\ref{eq:belief_update} \cite{ANDRIOTIS2019106483}.

In our MOPOMDP formulation, we stated that the \textit{probability of collapse} reward $\log{\left ( 1 - \risk_{t}\right )}$ as the inverse log probability of collapse occurring at timestep $t$. Using this inverse log probability is based on two key considerations. Firstly, using the log probability ensures that the episodic return of the probability of collapse $R_{\risk} = \sum_{t=0}^{H}\log{\left ( 1 - \risk_{t}\right )}$ is a valid probability value. Secondly, $R_{\risk}$ will be the log probability of no collapse occurring during the episode; therefore, we can use the base of the natural logarithm $e$ to obtain the probability of collapse over the whole episode easily: 
\begin{align}
    P_{\risk} &=1 - \prod_{t=0}^{H}\left (1 - \risk_{t} \right) \nonumber \\
      &= 1 -e^{\sum_{t=0}^{H}\log{\left ( 1 - \risk_{t}\right )}} \nonumber \\
      &=1 - e^{R_{\risk}}\label{eq:log_prob_one_collapes}.
\end{align}
\section{Multi-Objective Deep
Centralized Multi-Agent
Actor-Critic (MO-DCMAC)} \label{sec:modcmac}
\begin{figure*}[h]
    \centering
    \includegraphics[width=\textwidth]{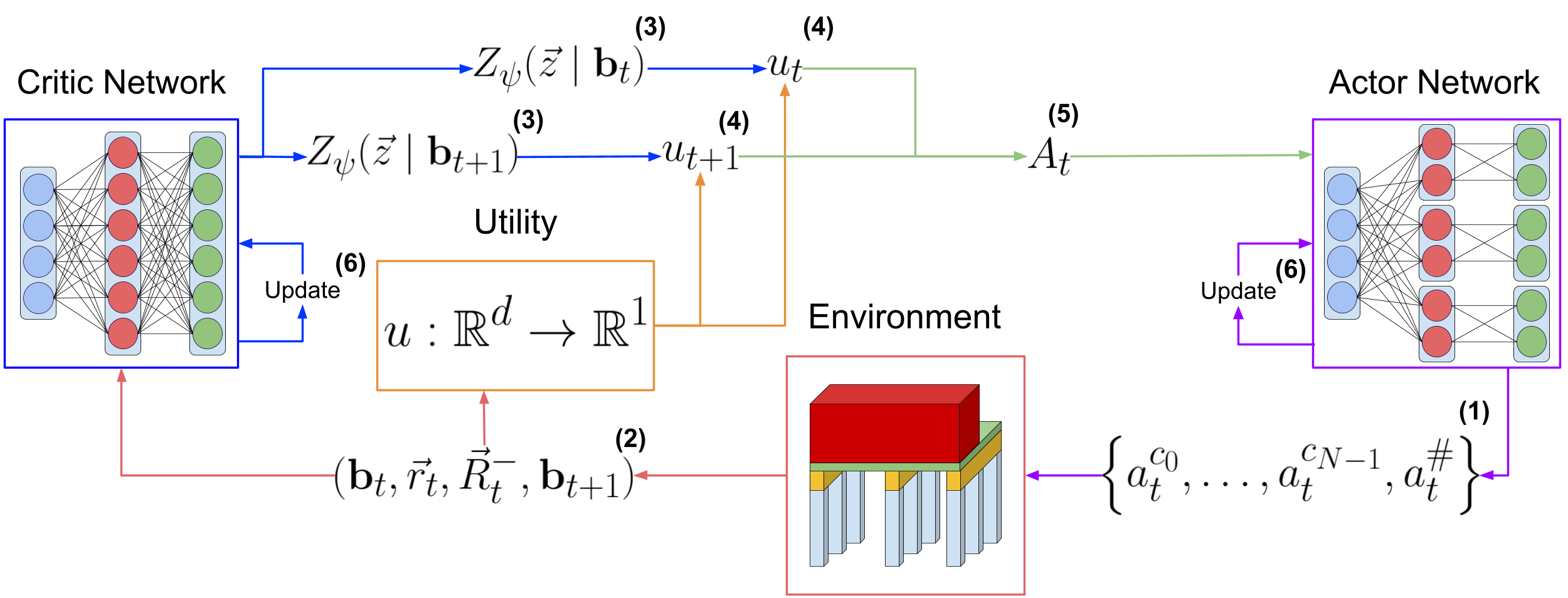}
    \caption{This figure illustrates the training sequence for MO-DCMAC, whereby the blue nodes are the input layers, the red hidden layers, and the green output layers. Algorithm~\ref{algo:mo-dcmac} is the pseudocode of MO-DCMAC and shows how each equation is used. Step (1) involves the actor sampling actions for each component and a collective global action from the current belief state $\mathbf{b}_t$. In step (2), these actions are processed by the environment to determine the subsequent belief state $\mathbf{b}_{t+1}$ and to generate the reward vector $\vec{r}_t$. This reward is then used with the accrued reward $\vec{R}^{-}_{t}$ for further calculations. Step (3) has the critic network predicting the joint distribution of the returns across objectives, denoted as $Z_{\psi}(\vec{z} \mid \mathbf{b}_t)$ for the current state and $Z_{\psi}(\vec{z} \mid \mathbf{b}_{t+1})$ for the next. In step (4), these predictions, along with the reward vector $\vec{r}_t$ and accrued reward $\vec{R}^{-}_{t}$, are used to compute the preference scores for the current $u_{t}$ and next state $u_{t+1}$. Step (5) employs these preference scores to calculate the advantage of the state $A_{t}$. Finally, step (6) updates the critic and actor. The overall loss combines actor and critic losses with an entropy bonus. Step (6) can be done at every step but can also be done at every $\eta$ step, whereby the actor and critic are updated from the previous $\eta$ steps.}
    \label{fig:MO-DCMAC_interaction}
\end{figure*}

\begin{algorithm}[h]
\caption{MO-DCMAC}\label{algo:mo-dcmac}
\begin{algorithmic}[1]
\Require $\eta$ Update Steps, $n_{\max}$ Number of Training Steps
\State Initialize actor weights $\theta$ and critic weights $\psi$.
\State Initialize step counter $n=0$.
\While {True}
\State Get $s_0$
\State $\textbf{b}_0 \gets b{\left(s_0 \right)}$
\For {$t \gets 1$ \textbf{to} $H$}
\State Select actions $a_t=\left \{ a^{\comp_0}_t, \dots, a^{\comp_{N-1}}_t, a^{\glob}_t \right \} \sim \pi_{\theta}(\cdot\mid\mathbf{b}_t, \frac{t}{H}, \vec{R}^{-}_{t})$.
\State Perform $a_{t}$ and receive $\vec{r}_t$ and observation $o_{t+1}$.
\State Calculate the new belief $\mathbf{b}_{t+1} \gets b(s_{t+1})$.
\State Store the tuple $((\mathbf{b}_t, \frac{t}{H}, \vec{R}_t^{-}), a_{t}, \vec{r}_t, (\mathbf{b}_{t+1}, \frac{t+1}{H}, \vec{R}_{t+1}^{-}))$
\State $n \gets n + 1$
\If {$n \mod \eta = 0$}
\State Get previous $\eta$ training examples.
\State Calculate advantage using Equation~\ref{eq:mocac_advantage}.
\State Calculate critic loss using Equations~\ref{eq:mocac_projection} and \ref{eq:mocac_critic_loss}.
\State Calculate policy loss with Equation~\ref{eq:actor_loss_modcmac}.
\State Calculate final loss with Equation~\ref{eq:MODCMAC_LOSS_COMB}.
\State Reset the buffer.
\EndIf
\If{$n \geq n_{\max}$}
\State \Return
\EndIf
\EndFor
\EndWhile
\end{algorithmic}
\end{algorithm}

In the previous section, Section~\ref{sec:maintenance_problem}, we discussed the formulation of the maintenance optimization problem, for which we want to learn a maintenance strategy and how we formalized this as a MOPOMDP. However, standard RL methods face significant challenges in learning an optimal policy, primarily due to two essential reasons.

The first is the curse of dimensionality in the action space. The action space grows exponentially to the number of components and possible maintenance actions; for example,  with 12 components and three possible actions, the action space would be $3^{12}=531441$. Andriotis and Papakonstantinou introduced \textit{Deep Centralized Multi-Agent Actor-Critic} (DCMAC) 
\cite{ANDRIOTIS2019106483}, an actor-critic method, to alleviate this. DCMAC uses a factorized action space representation whereby the actor network has an output for each component that outputs an action for that component. 

Secondly, our maintenance optimization problem is a multi-objective problem with a known non-linear utility function. Moreover, we want to learn a policy using the ESR criterion (Equation~\ref{eq:esr_criterion}); however, this introduces significant complexity because if the utility function is non-linear, then the Bellman equation becomes invalidated (Equation~\ref{eq:esr_bellman_invalid}) \cite{Hayes2022}. Reymond et al. introduced \textit{Multi-Objective Categorical Actor-Critic} (MOCAC) \cite{Reymond2023} that utilizes a distributional critic, which learns a multi-variate distribution over the returns of the objective. MO-DCMAC requires this multi-variate distribution over the return, introduced by Reymond et al. \cite{Reymond2023}, to be able to learn an optimal policy with the ESR criterion in a multi-objective problem setting. We explain this further in this section.

Our proposed method, \textit{Multi-Objective Deep Centralized Multi-Agent Actor-Critic (MO-DCMAC)}, combines DCMAC \cite{ANDRIOTIS2019106483} and MOCAC \cite{Reymond2023}. Combining these methods allows us to overcome the curse of dimensionality of action space by using the actor network of DCMAC and the ability to directly optimize over a non-linear utility function with the ESR criterion by using the critic network of MOCAC.

The goal of MO-DCMAC is to learn a policy that returns the highest utility for a given multi-objective maintenance optimization problem with a known non-linear utility function under the ESR criterion. We needed to integrate DCMAC \cite{ANDRIOTIS2019106483} and MOCAC \cite{Reymond2023} into MO-DCMAC to achieve this goal. The critic of MO-DCMAC learns the multi-variate distribution of the objectives returns instead of the expected return because to learn under the ESR criterion; we require the return as input for the utility. Moreover, this allows MO-DCMAC to correctly calculate the advantage $A_{t}$, which is used to update the actor. The interaction between the critic and actor can be seen in Figure~\ref{fig:MO-DCMAC_interaction} and Algorithm~\ref{algo:mo-dcmac}. The exact implementation details will be explained in the remainder of this section.

The critic and most of the loss calculations in MO-DCMAC are based on previous work introduced by Reymond et al.~\cite{Reymond2023}. Reymond et al. introduce a distributional critic that learns the multi-variate distribution $Z$ of the return of the objectives. They based this on the work of Bellemare et al.~\cite{DBLP:journals/corr/BellemareDM17}, whereby the major difference is that Reymond et al.~\cite{Reymond2023} learn a multi-variate distribution. The reason why the critic learns the distribution of the objectives instead of the expected vectorial return is that when the expected vectorial return is scalarized by a utility function $u$, it would learn under the SER criterion because the vectorial value-function learns the expected return (Equation~\ref{eq:value_function}). To ensure that MO-DCMAC can learn an optimal policy under the ESR criterion, episodic return is required instead~\cite{Reymond2023}.

Therefore, we utilize the critic introduced by Reymond et al.~\cite{Reymond2023} to learn a distribution $Z$ through the following equation:
\begin{equation}\label{eq:critic_mocac}
    V{\left (s_t \right)} = \sum_{i \in \left [\left \{0,\dots,\mathcal{N}-1  \right \} \right ]^d}\vec{z}_{i}Z_{\psi}{\left(\vec{z}_{i} \mid s_{t} \right)}.
\end{equation}
where $\vec{z}_{i}$ is the support atom for the vectorial return. Each atom $\vec{z}_{i}$ for $\forall i \in \left [\left \{0,\dots,\mathcal{N}-1  \right \} \right ]^d$ is calculated by,
\begin{align} 
        &\vec{z}_{i} = \vec{V}_{\min} + i\Delta\vec{z} \nonumber \\
        &\text{with}\: \Delta\vec{z}:= \frac{\vec{V}_{\max} - \vec{V}_{\min}}{\mathcal{N}- 1}\label{eq:atoms_mocac}
\end{align}
This support atom is calculated by providing $V_{\min}, V_{\max} \in \mathbb{R}$ for each objective, which represents the minimum and maximum return for that objective. Therefore, $\vec{z}$ represents the possible returns we can have for the objectives. The critic $Z_{\psi}{\left(\vec{z}_{i} \mid s_{t} \right)}$ learns the probability that any given return in $\vec{z}$ will occur from the current state $s_{t}$. $\vec{z}$ itself consists of $\mathcal{N}^{d}$ atoms, where each atom represents a possible return of the objectives. The number of atoms $\mathcal{N}$ can be chosen arbitrarily; however, Reymond et al.~\cite{Reymond2023} assume the same number of atoms for each objective. 

Due to Equations~\ref{eq:critic_mocac} and \ref{eq:atoms_mocac} introduced by Reymond et al.~\cite{Reymond2023}, the critic learns the discretized distribution of the objectives instead of the expected return. Because $\vec{z}$ does not represent the expected return but is defined as a return, MO-DCMAC can learn under the ESR criterion. Therefore, in combination with the accrued reward $\vec{R}^{-}_{t}=\sum_{k=0}^{t-1}\gamma^{k}\vec{r}_{k}$, we can calculate a preference score $u_{t}$ \cite{Reymond2023} under the ESR criterion by:
\begin{equation} \label{eq:utility_preference}
    u_{t} = \sum_{i \in \left [\left \{0,\dots,\mathcal{N}-1  \right \} \right ]^d}u{\left (\vec{R}^{-}_{t} +\vec{z}_{i} \right )} Z_{\psi}{\left(\vec{z}_{i} \mid s_{t} \right)}.
\end{equation}

With the preference score $u_{t}$ being calculated through Equation~\ref{eq:utility_preference}, the advantage $A(s_t, a_t)$ of MO-DCMAC is calculated through Equation \ref{eq:mocac_advantage}~\cite{Reymond2023}
\begin{align}
    A{\left (s_{t}, a_{t} \right)}=& \Bigg ( \sum_{i \in \left [\left \{0,\dots,\mathcal{N}-1  \right \} \right ]^d} u\left (\vec{R}^{-}_{t} + \gamma^{t} \left(\vec{r}_{t} + \gamma \vec{z}_i\right) \right)\nonumber \\ 
    &\cdot Z_{\psi}{\left (\vec{z}_{i}\mid s_{t+1} \right )} \Bigg )\nonumber \\
    &-\Bigg (\sum_{i \in \left [\left \{0,\dots,\mathcal{N}-1  \right \} \right ]^d} u\left ( \vec{R}^{-}_{t} + \gamma^{t}\vec{z}_{i}\right ) \nonumber \\ 
    &\cdot Z_{\psi}{\left (\vec{z}_{i}\mid s_{t} \right )}\Bigg )\label{eq:mocac_advantage}
\end{align}
Reymond et al.~\cite{Reymond2023} did not originally normalize the advantage. However, we found that normalizing significantly increased training stability; therefore, the final advantage used is:
\begin{equation}\label{eq:norm_adv}
    \hat{A}{(a_{t}, s_{t})} = \frac{A{(a_{t}, s_{t})} - \mu_{A}}{\sigma_{A}},
\end{equation}
where $\mu_A$ is the mean advantage of the current batch and $\sigma_A$ is the standard deviation of the advantage of the current batch.

For the critic's loss calculation, first the distributional Bellman update $\hat{\mathcal{T}}\vec{z}_{j}=\vec{r}_{t} + \gamma z_{j}$ for each atom $\vec{z}_{j}$ is calculated. Reymond et al. \cite{Reymond2023} then distributes the probability of $Z_{\psi}{\left( \vec{z}_{i} \mid s_{t+1}\right)}$ to the immediate neighbors of $\hat{\mathcal{T}}\vec{z}_{j}$, whereby the projected update is:
\begin{align} 
    \Phi \hat{\mathcal{T}}Z_{\psi}\left (s_{t} \right) =& \Bigg ( \sum_{j} \left [ 1-\frac{\left | \left [\hat{\mathcal{T}}\vec{z}_j \right]^{V_{\max}}_{V_{\min}} - \vec{z}_{i}\right|}{\Delta z}\right ]_{0}^{1}\nonumber \\ 
    &\cdot Z_{\psi}{\left( \vec{z}_{i} \mid s_{t+1}\right)} \Bigg ), \nonumber \\&\forall i \in \left [\left \{0,\dots,\mathcal{N}-1 \right \} \right ]^{d}\label{eq:mocac_projection},
\end{align}
with $[\cdot]^{a}_{b}$ meaning that it is bounded between $a$ and $b$.

The loss of the critic is then calculated with Equation~\ref{eq:mocac_projection} through the cross-entropy term of the KL-divergence:
\begin{equation} \label{eq:mocac_critic_loss}
    \mathcal{L}^{V}(V)=D_{KL}\left (\Phi \hat{\mathcal{T}}Z{\left (s_{t} \right)} \parallel Z{\left(s \right)} \right)
\end{equation}

Previously, we stated that we based the actor architecture on DCMAC \cite{ANDRIOTIS2019106483}. The main contribution of DCMAC is the factorized action output, where for each component $\comp_{i}$, there is a distinct output layer and, depending on the setup, one or more hidden layers. 

DCMAC is an actor-critic method, which calculates the loss for the actor similarly to Equation~\ref{eq:actor_critic_pol_update}. However, the main difference is that DCMAC has to combine the output of all the output heads in the actor network for the loss calculation, yielding:
\begin{equation} \label{eq:actor_loss_dcmac}
    \mathcal{L}^{\pi}{(\pi)}=-w_{t}\left (\sum_{j=0}^{N-1}\log{\left(\pi^{\comp_{j}}_{\theta}{\left ( a^{\comp_{j}}_{t} \mid \mathbf{b}_t\right)}\right)} \right)A{\left(\mathbf{b}_{t}, a_{t} \right)},
\end{equation}
where $w_{t}=\min\left(q, \frac{\pi\left(a_{t} \mid s_{t} \right)}{\mu\left(a_{t} \mid s_{t} \right)} \right)$ is a truncated importance sampling \cite{truncated_importance} to address the high variance that log summation may introduce, $c_{j}$ the component for which an action is selected, and $\mathbf{b}_{t}$ is the belief at timestep $t$ because DCMAC is used to find an optimal policy in a POMDP. The reason a truncated importance sampling is needed is that DCMAC is an off-policy method because it collects training examples in a replay buffer and samples during training from it. In contrast, MO-DCMAC is an on-policy method, meaning we do not reuse training examples similar to MOCAC \cite{Reymond2023}. Therefore, the importance sampling is not necessary anymore due to $\frac{\pi\left(a_{t} \mid s_{t} \right)}{\pi\left(a_{t} \mid s_{t} \right)}=1$. Additionally, we introduce another separate head that selects a global action head from $A^{\glob}$. Therefore, the modified policy loss calculation will be:
\begin{align} 
    \mathcal{L}^{\pi}{(\pi)}=-\bigg(&\sum_{j=0}^{N-1}\log{\left(\pi^{\comp_{j}}_{\theta}{\left ( a^{\comp_{j}}_{t} \mid \mathbf{b}_t\right)}\right)} \nonumber \\ &+ \log{\left(\pi^{\glob}_{\theta}{\left ( a^{\glob}_{t} \mid \mathbf{b}_t\right)}\right)} \bigg)\hat{A}{\left(\mathbf{b}_{t}, a_{t} \right)}\label{eq:actor_loss_modcmac},
\end{align}
where $\hat{A}(s_{t}, a_{t})$ is calculated through Equations~\ref{eq:mocac_advantage} and \ref{eq:norm_adv}. Similar to MOCAC \cite{Reymond2023}, we combine the critic loss and the policy loss through,
\begin{equation} \label{eq:MODCMAC_LOSS_COMB}
    \mathcal{L}\left (\pi, V \right) = \mathcal{L}^{\pi}{\left(\pi \right)} + \lambda_{\text{val}} \mathcal{L}^{V}(V) - \lambda_{\text{ent}} \mathcal{H}(\pi),
\end{equation}
where $\mathcal{H}(\pi)$ is the entropy bonus from $\pi$, $\lambda_{\text{val}}$ the coefficient for the value loss and $\lambda_{\text{ent}}$ the coefficient for the entropy bonus. Both $\lambda_{\text{value}}$ and $\lambda_{\text{ent}}$ are hyperparameters and control the contribution of the value and entropy loss, respectively.

MO-DCMAC takes three input values, namely $\mathbf{b}_{t}, \frac{i}{H}$ and $ \vec{R}^{-}_{t} $. The first value is the belief $\mathbf{b}_{t}$ at timestep $t$, which we update through Equation~\ref{eq:belief_update}, $\frac{t}{H}$ is the normalized timestep, with $H$ being the horizon of the episode. Lastly, we included the accrued reward $\vec{R}^{-}_{t}$ as necessitated by the multi-objective setting under the ESR with a non-linear utility \cite{Reymond2023}.

\section{Experimental Setup}\label{sec:exp_setup}
This section explains the details of the experiments we conduct to show the capabilities of our introduced method, MO-DCMAC, to minimize the cost and probability of collapse through a given utility function that weighs these two objectives. We start by introducing the environments we used in our experiments. Afterward, we explain the utility functions used to train MO-DCMAC and state the baseline policies we used to evaluate the performance of MO-DCMAC.
\subsection{Environments}
We test the performance of MO-DCMAC in three different maintenance environments, namely: (i) Simple Homogeneous Asset, in which components are of the same type, and the calculation of the probability of collapse has no dependencies. (ii) Quay Wall Asset environment has multiple types of components and dependencies. This environment is based on the inner-city quay wall of Amsterdam. (iii) Larger Quay Wall Asset, a larger version of the above environment, which has more components and dependencies.

\subsubsection{Simple Homogeneous Asset} \label{sec:experiments:simple_homogeneous_asset}
In the Simple Homogenenous Asset environment, maintenance is planned for an asset with eight components of the same type and has the same degradation rate. The effect of a failing component is also independent of other failing components, meaning that there are eight sub-dependency groups because each sub-dependency group has one component. The full details of this environment can be found in Appendix~\ref{appendix:env:simple}.

\subsubsection{Quay Wall Asset} \label{sec:smaller_quay_wall}
\begin{figure}[h]
    \centering
    \includegraphics[width=0.4\textwidth]{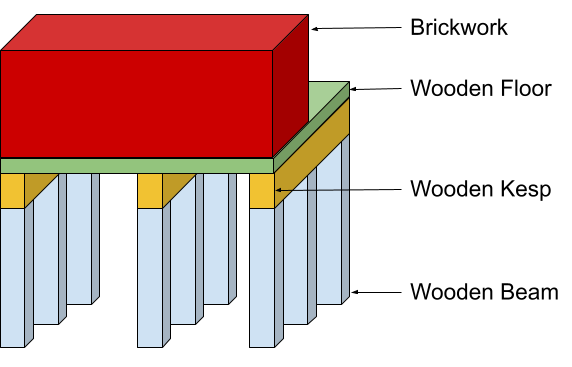}
    \caption{Representation of the Quay Wall Asset environment. Maintenance will only be done for the wooden components.}
    \label{fig:cross_section_qual_wall}
\end{figure}

The second environment is the Quay Wall Asset environment. This environment is based on the historical inner-city quay walls of Amsterdam\footnote{For an overview of what is currently being done in terms of maintenance, see \url{https://www.amsterdam.nl/projecten/kademuren/} (in Dutch)}. \cite{grimburgwal}. Figure~\ref{fig:cross_section_qual_wall} shows the components of this environment. In this environment, we only plan the maintenance of the wooden components because these components deteriorate significantly faster than the brickwork and are more critical for structural integrity. We considered three types of components: Wooden Beams, Wooden Kesps, which connect to the Wooden Beams, and the Wooden Floor, which rests on the Wooden Kesps. In this environment, we considered dependencies. For instance, the Wooden Beams connected to the same Wooden Kesp belong to the same sub-dependency group. The contribution to the probability of collapse is almost negligible if one Wooden Beam is in the worst state; however, if two or more are, the contribution to the probability of collapse is significantly higher due to the less support this group of Wooden Beams can give to the whole quay wall. More information, such as the degradation matrices and the specific groups, are found in Appendix~\ref{appendix:env:normal}.

\subsubsection{Larger Quay Wall Asset} \label{sec:larger_quay_wall}
The last environment we consider is the Larger Quay Wall Asset. This environment is almost identical to the previous one, except that this environment doubles the total components. We use this environment to test how well MO-DCMAC scales to the number of components. The exact specifications of this environment can be found in Appendix~\ref{appendix:env:difficult}.

\subsection{Utility Functions} 
We test each of the environments with two different utility functions: a Threshold utility function, where the probability of collapse should always be below a certain threshold; otherwise, a penalty will be received, and a utility function based on the Failure Mode, Effects, and Criticality Analysis (FMECA) methodology, which asset managers use to determine the desired effect of the maintenance strategy.

As previously discussed, the input for the utility functions is the episodic return of cost $R_{\cost} = \sum_{t=0}^{H} \gamma^{t}(-\cost_{t})$ and the log probability of collapse $R_{\risk}=\sum_{t=0}^{H} \gamma^{t}\log(1 -\risk_{t})$, where the log probability of collapse $R_{\risk}$ is calculated to $P_{\risk}$ (Equation~\ref{eq:log_prob_one_collapes}). This probability value $P_{\risk}$ is the probability that during the planned period, a collapse can occur.

In our results, we examine how these utilities lead to different policies by comparing the differences in the objective scores, cost $\cost$, and probability of collapse $\risk$.
\subsubsection{Threshold Utility} \label{sec:uti_threshold}
The Threshold utility function is based on standard safety requirements set for infrastructural assets, namely that an asset should adhere to a minimal safety \cite{safety_requirements}. With the Threshold utility, MO-DCMAC should learn a policy that optimizes for cost while ensuring the probability of collapse will not exceed a certain threshold, such that the asset adheres to the minimal safety requirements. Minimizing the probability of collapse beyond the threshold will not improve the utility score. Therefore, MO-DCMAC should find an optimal policy that ensures the proposed maintenance plan never exceeds the threshold but does not drastically increase cost. We formulated the utility as follows:
\begin{equation}\label{eq:threshold_utility}
    u(R_{\cost}, P_{\risk}) = \begin{cases}
R_{\cost} & \text{ if } P_{\risk} \leq 0.1 \\ 
3(R_{\cost} + 1) & \text{ if } 0.1 < P_{\risk} \leq 0.2 \\ 
5(R_{\cost} + 2) & \text{ otherwise } 
\end{cases}.
\end{equation}
In our experiments, a penalty is received if $P_{\risk} > 0.1$, which is increased if it is $P_{\risk}>0.2$. However, these thresholds can be higher or lower, or the penalty amount can be changed depending on the asset, with the only restriction that the utility should be monotonically increasing.

In our experiments, we show the effect of the discount factor $\gamma$ on achieving an optimal policy with Threshold utility; namely, if the discount factor is too low, it could result in a learned policy that believes it is below the threshold because the discounted return of the probability of collapse is. Still, it is not because the non-discounted return is above the threshold. 

This difference between the non-discounted and discounted returns is not a problem for standard RL. However, for utility-based MORL, this difference could be because we optimize over the utility function. This is especially essential for utility functions that can contain a strict threshold for an objective and where optimizing the objective is not beneficial for the utility score if the threshold has been met, such as the Threshold utility. Therefore, we will test the following discount factors: $0.9$, $0.975$, $0.99$, $0.995$, and $1$. These experiments are conducted with the \textit{Simple Homogeneous Asset} environment (Section \ref{sec:experiments:simple_homogeneous_asset}).

\subsubsection{FMECA Utility}

The FMECA utility function is based on the \textit{Failure Mode Effect and Criticality Analysis (FMECA)} methodology~\cite{borgovini1993failure}. The goal of FMECA, in a maintenance planning setting, is to determine the most optimal maintenance strategy that limits the probability that a failure mode occurs while ensuring that other objectives are also optimized. Within FMECA, a wide range of objectives can be used. In our setting, we will focus only on our existing objectives: probability of collapse and cost; however, this utility can be extended to include other objectives. 

\begin{figure}[h]
    \centering
    \includegraphics[width=0.45\textwidth]{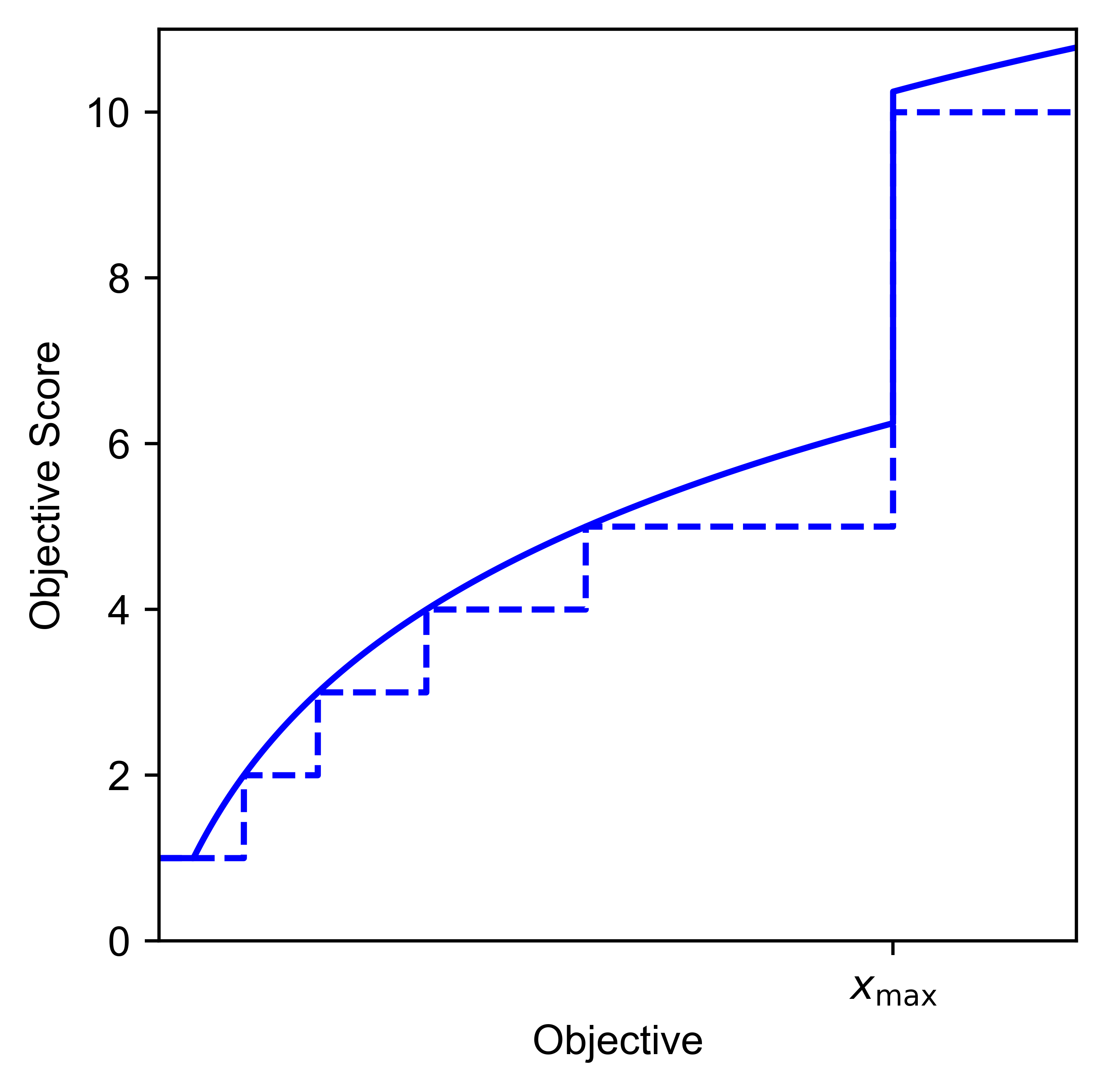}
    \caption{The figure shows how we smoothed out the objective scoring for the FMECA utility. The dashed line is the normal, binned scoring from the original FMECA, and the solid line is the scoring we used to improve the training of MO-DCMAC. The increase from 5 to 10 is done at $x_{\max}$.}
    \label{fig:fmeca_score_bining}
\end{figure}

The FMECA methodology we use in this paper is based on the one used by asset managers in Amsterdam for the inner-city quay walls. These asset managers give a score to each objective based on the effect of a proposed maintenance strategy on that objective. Each objective is classified into one of six bins to compute its utility score. The first five bins incrementally assign scores from one to five, while the last bin gives a score of ten; lower scores are preferable. Notably, each bin's range decreases logarithmically from the worst to the best bin. However, this binning approach can potentially prolong MO-DCMAC's training time. It may either trap MO-DCMAC in a local optimum, taking longer to escape, or, in the worst-case scenario, prevent escaping altogether. Therefore, we used a logarithm for smoother score growth to mitigate the risk of this happening (Figure~\ref{fig:fmeca_score_bining}). We formulate the FMECA utility as follows:
\begin{align}
       c_{\text{utility}}(R_{\cost}, C_{\max}) =&
        6 \log_{10}{\left (1 + 10\frac{R_{\cost}}{C_{\max}}\right)}\notag\\ 
        &+ \text{Pen}(R_{\cost}, C_{\max}) \label{eq:cost_uti}\\
    f_{\text{utility}}(P_{\risk}, F_{\max}) =& 6 \log_{10}{\left (1 + 10\frac{P_{\risk}}{F_{\max}}\right)} \notag \\
    &+ \text{Pen}(P_{\risk}, F_{\max}) \label{eq:collapse_uti} \\
    \text{where } \text{Pen}(x, x_{\max})&=  \begin{cases}
         4 & \text{ if } x \geq x_{\max} \\ 
         0 & \text{ otherwise} 
     \end{cases} \label{eq:fmeca_pen}
\end{align}
Equations~\ref{eq:cost_uti}, \ref{eq:collapse_uti}, and \ref{eq:fmeca_pen} are used to calculate the FMECA score of each objective. We employed logarithm to base 10 ($\log_{10}$) to match the original FMECA scoring, where binning was used to determine the score and where the bins range decreased logarithmically. This logarithmic scoring alone is sufficient to simulate the first five bins that score from one to five. However, we had to add Equation~\ref{eq:fmeca_pen} to achieve the correct behavior related to the last bin, where a score of ten is assigned. This penalty is added if either the cost $R_{\cost}$ is higher than $C_{\max}$ or the probability of collapse is higher than $F_{\max}$, such that the objective score will be increased from six to ten. For our experiments, we set $F_{\max}=0.2$ and $C_{\max}=4$. However, lower or higher values are possible because $F_{\max}$ and $C_{\max}$ act as weights within the FMECA utility.

\begin{align} 
    u = -(&\max{\left( 1, c_{\text{utility}}(R_{\cost}, C_{\max})\right )} \nonumber \\ &\cdot  \max{\left( 1, f_{\text{utility}}(P_{\risk}, F_{\max}) \right )}) \label{eq:final_uti}
\end{align}

Equation~\ref{eq:final_uti} calculates the last step of the utility and combines the score calculated through Equations~\ref{eq:cost_uti} and \ref{eq:collapse_uti}. We make the utility negative because, within the FMECA method, a lower FMECA score is preferred. We also ensure that $u$ never can be lower than 1, which is the minimum score in FMECA. Moreover, if we would not set both $c_{\text{utility}}$ and $f_{\text{utility}}$ to at least 1, MO-DCMAC would likely learn to do no maintenance at all because when the cost is 0, the utility score would also be 0.
\subsection{Baseline Policies}
In our experiments, we will compare MO-DCMAC to baseline policies based on current maintenance strategies. In this subsection, we will explain each of these baseline policies. 
\begin{itemize}
    \item Year-Based Action (YBA) policy. This policy does no inspection and will take a \emph{repair} or \emph{replace} action for each component on year interval.
    \item Year-Based Inspection and Condition-Based Action (YBI-CBA) policy. The YBI-CBA policy inspects at a set interval whereby maintenance action is taken at the next time step based on the inspection.
    \item Condition-Based Inspection and Condition-Based Action (CBI-CBA) policy. The CBI-CBA policy is similar to the YBI-CBA policy; however, the inspection is not done at an interval but when a percentage of the components are in a worse state. A component is in a worse state if it is in one of the three worst states. Without an inspection, we can only detect if a component is in the first two or the last three states, but we cannot determine the exact state.
\end{itemize}
The CBA policies are set to \textit{do nothing} if a component is in the first state, \textit{repair} if in the second or third state, and \textit{replace} if in the fourth or last state. We optimized the CBI, YBI, and YBA policies by either finding the optimal yearly interval at which an inspection should be done for YBI or repairing or replacing should be done for YBA. We optimized CBI for the percentage of components that should be in a worse state before an inspection is done. Our results will only show the performing policy for each environment and utility.

We did not include other single-objective RL methods or multi-objective RL methods since they cannot learn optimal policies for the given problem setting. Reymond et al. \cite{Reymond2023} showed that single-objective RL methods trained on only the utility score, given at the last state, result in highly unstable policies that always underperform. Other multi-objective RL methods cannot learn with the ESR criterion \cite{Hayes2022}, and therefore, we cannot compare the results of MO-DCMAC with the results of these methods. We did not include MOCAC \cite{Reymond2023} since it would require substantial adjustments for it to be able to function in our problem setting. Lastly, other methods for utility-based maintenance optimization, such as genetic algorithms \cite{AllahBukhsh2020, earthquake_ga, GUAN2022130103}, cannot scale to problem instances of this size.

\subsection{Hyperparameters}
In Table~\ref{tab:hyperparameters}, we state the hyperparameters we used for our experiments. The hyperparameters were found by manual search. Most importantly, we only adjusted the hyperparameters between the utility function, and only $\gamma$, $V_{\min}$, and $V_{\max}$. This entails that we mostly use the same hyperparameters between the environments. The network architecture is a feedforward neural network. The actor network has one shared input layer. Each head has a hidden layer of size 50 and an output layer. The critic network consists of one hidden layer with 150 neurons.

For all the environments, we set the cost as the percentual value of replacement. Therefore, if all the components are replaced simultaneously, the cost for that time step would be 1. We found that during the initial testing, this significantly improved the accuracy because the original value for the cost of components was in the thousands, and higher values are harder to learn for a neural network. Another benefit is that this allowed hyperparameters to be shared between the environments. Otherwise, the $V_{\min}$ for cost must differ between environments, which could also affect other hyperparameters. Lastly, an episode consists of 50 years, each timestep representing a single year.
\begin{table}[h]
\caption{The hyperparameters used for the experiments}\label{tab:hyperparameters}%
\begin{tabular}{@{}lll@{}}
\toprule
Utility & Threshold  & FMECA  \\
\midrule
Training Steps & $2.5 \times 10^7$   & $2.5 \times 10^7$   \\
Actor LR    & (2E-4, 2E-5)\footnotemark[1]   & (2E-4, 2E-5)\footnotemark[1]  \\
Critic LR    & (2E-3, 2E-4)\footnotemark[1]   & (2E-3, 2E-4)\footnotemark[1] \\
Activation & Tanh & Tanh \\
Clip Grad Norm & 100   & 100  \\
Update Every & 128 & 128 \\
$\mathcal{N}$ & 11 & 11 \\
$\lambda_{\text{val}}$ & 0.5 & 0.5 \\
$\lambda_{\text{ent}}$ & 0.001 & 0.001 \\
$\gamma$ & 0.995 & 0.975 \\
$V_{\min}$ Cost & -12 & -4 \\
$V_{\max}$ Cost & 0 & 0 \\
\begin{tabular}[c]{@{}l@{}}$V_{\min}$ Log Probability \\ of Collapse\end{tabular} & -0.1 & -0.02 \\
\begin{tabular}[c]{@{}l@{}}$V_{\max}$ Log Probability \\ of Collapse\end{tabular} & 0 & 0 \\

\botrule
\end{tabular}
\footnotetext[1]{The learning rate is linearly annealed during training from the start value to the end value.}
\end{table}

\subsection{Software and Hardware}
We used Python 3.10 for our implementation. We used PyTorch 2.1, and Gymnasium 0.28 for the environments. The experiments were conducted on an Apple M1 Pro CPU with 16GB RAM. 

\section{Results and Discusion}
\begin{table*}[h]
\caption{The results of the experiments in which we note the mean and standard deviation of 5000 episodes. The score column contains the utility score of the corresponding utility function. The bold results are the best results for the environment and utility combination, whereby if the mean is equal, a lower standard deviation is preferable. For each column, a lower score is preferable.}\label{table:all_results}%
\begin{tabular}{@{}llllll@{}}
\toprule
Environment & \begin{tabular}[c]{@{}l@{}}Utility\\ Score\end{tabular} & Policy & Score & Cost & \begin{tabular}[c]{@{}l@{}}Probability\\ of Collapse\end{tabular} \\
\midrule
 Simple& FMECA & YBA-Repair & 36.104 ± 5.74 & \textbf{0.75 ± 0} & 0.747 ± 0.224 \\
Homogenous & & YBA-Replace & 41.394 ± 41.637 & 10 ± 0 & 0.067 ± 0.098 \\
Asset & & CBI-CBA & 13.271 ± 7.777 & 1.623 ± 0.338 & \textbf{0.056 ± 0.052} \\
 & & YBI-CBA & 13.53 ± 9.339 & 1.636 ± 0.319 & 0.058 ± 0.061 \\
 & & MO-DCMAC & \textbf{13.057 ± 7.419} & 1.471 ± 0.323 & 0.059 ± 0.052 \\
 & Threshold & YBA-Repair & 13.258 ± 2.412 & \textbf{0.75 ± 0} & 0.749 ± 0.23 \\
 & & YBA-Replace & 17.355 ± 13.878 & 5 ± 0 & 0.158 ± 0.177 \\
 & & CBI-CBA & 2.451 ± 2.467 & 1.615 ± 0.339 & 0.054 ± 0.052 \\
 & & YBI-CBA & 2.58 ± 3.181 & 1.651 ± 0.304 & \textbf{0.053 ± 0.057} \\
 & & MO-DCMAC & \textbf{2.292 ± 2.455} & 1.498 ± 0.325 & 0.058 ± 0.051 \\
 Quay Wall & FMECA & YBA-Repair & 16.486 ± 15.198 & 2.5 ± 0 & 0.063 ± 0.096 \\
Asset & & YBA-Replace & 17.451 ± 14.79 & 10 ± 0 & 0.01 ± 0.031 \\
 & & CBI-CBA & 5.566 ± 3.544 & 1.49 ± 0.227 & \textbf{0.01 ± 0.021} \\
 & & YBI-CBA & 9.567 ± 9.54 & 1.48 ± 0.2 & 0.039 ± 0.08 \\
 & & MO-DCMAC & \textbf{5.072 ± 3.158} & \textbf{1.028 ± 0.156} & 0.015 ± 0.02 \\
 & Threshold & YBA-Repair & 5.404 ± 6.073 & 2.5 ± 0 & 0.061 ± 0.089 \\
 & & YBA-Replace & 7.032 ± 6.534 & 5 ± 0 & 0.032 ± 0.075 \\
 & & CBI-CBA & 1.505 ± 0.492 & 1.479 ± 0.236 & \textbf{0.01 ± 0.018} \\
 & & YBI-CBA & 2.814 ± 3.888 & 1.489 ± 0.202 & 0.04 ± 0.082 \\
 & & MO-DCMAC & \textbf{1.133 ± 1.186} & \textbf{0.931 ± 0.15} & 0.028 ± 0.033 \\
 Larger Quay & FMECA &YBA-Repair & 29.127 ± 20.424 & 2.5 ± 0 & 0.14 ± 0.137 \\
Wall Asset & & YBA-Replace & 22.045 ± 21.016 & 10 ± 0 & \textbf{0.02 ± 0.045} \\
 & & CBI-CBA & 10.806 ± 7.183 & 1.453 ± 0.169 & 0.046 ± 0.052 \\
 & & YBI-CBA & 16.039 ± 13.19 & 1.541 ± 0.141 & 0.083 ± 0.108 \\
 & & MO-DCMAC & \textbf{7.293 ± 4.74} & \textbf{1.44 ± 0.284} & 0.022 ± 0.028 \\
 & Threshold & YBA-Repair & 10.1 ± 8.523 & 2.5 ± 0 & 0.143 ± 0.139 \\
 & & YBA-Replace & 10.576 ± 10.413 & 6 ± 0 & 0.057 ± 0.1 \\
 & & CBI-CBA & \textbf{2.214 ± 2.168} & \textbf{1.461 ± 0.169} & \textbf{0.044 ± 0.044} \\
 & &YBI-CBA & 4.583 ± 5.52 & 1.536 ± 0.144 & 0.084 ± 0.111 \\
 & &MO-DCMAC & 3.264 ± 4.114 & 1.568 ± 0.183 & 0.059 ± 0.07 \\
\botrule
\end{tabular}
\end{table*}
\begin{figure*}[h]
    \centering
    \includegraphics[width=\textwidth]{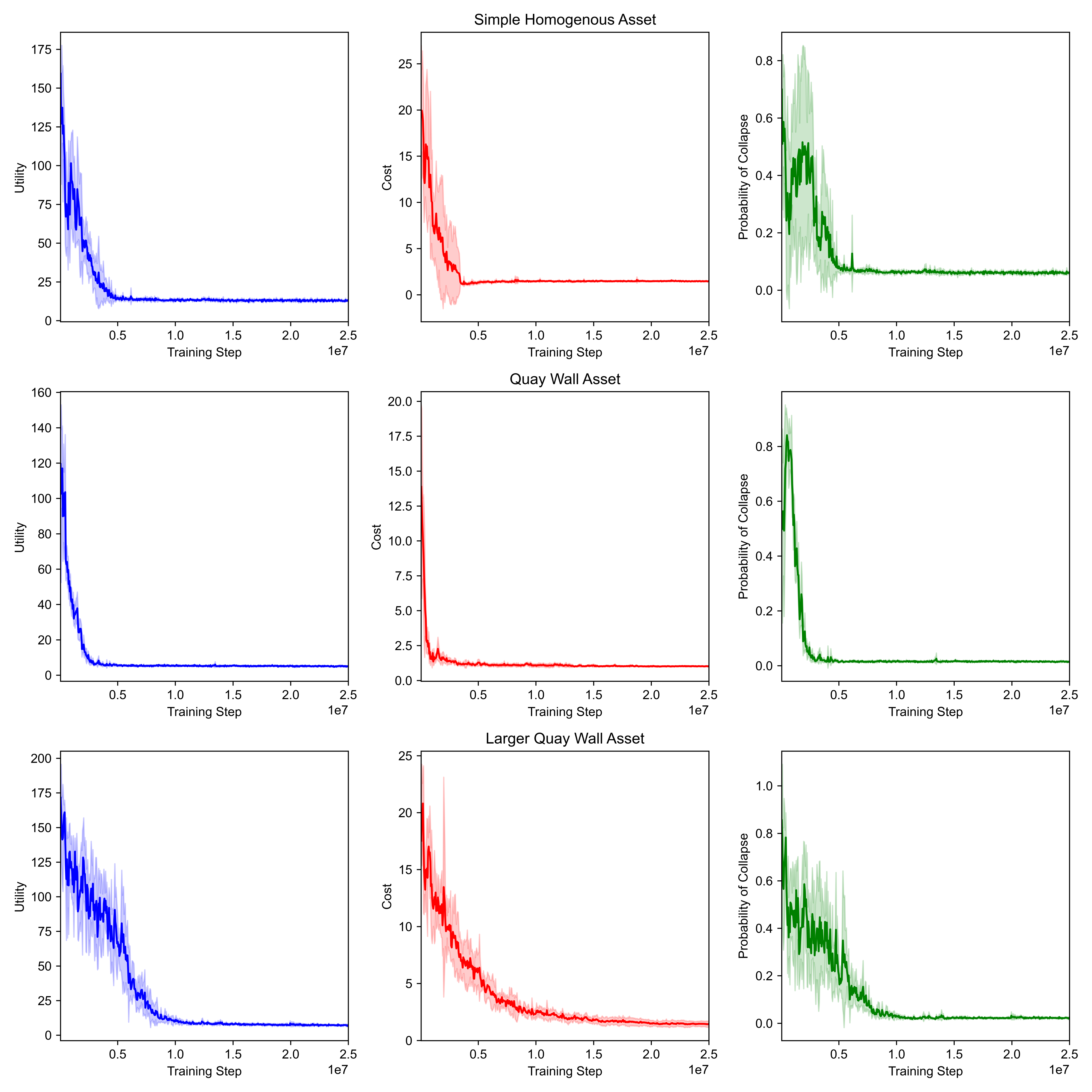}
    \caption{The training plots of MO-DCMAC with the FMECA utility function. The line shows the mean over 5 runs, with the shaded area being the standard deviation.}
    \label{fig:fmeca_training_plots}
\end{figure*}
\begin{figure*}[h]
    \centering
    \includegraphics[width=\textwidth]{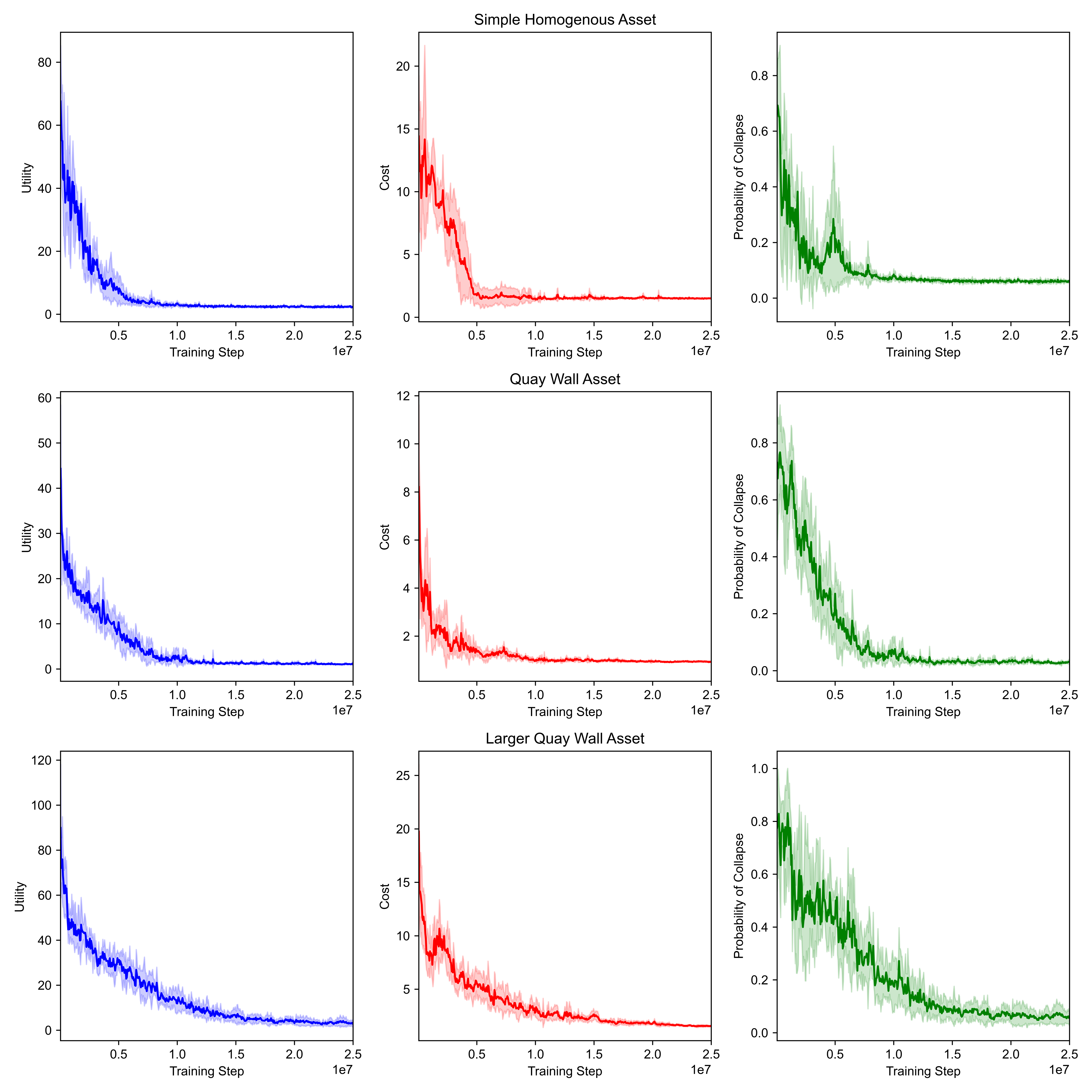}
    \caption{The training plots of MO-DCMAC with the Threshold utility function. The line shows the mean over 5 runs, with the shaded area being the standard deviation.}
    \label{fig:threshold_training_plots}
\end{figure*}
In Table~\ref{table:all_results}, we show the results of each environment for both utility functions, while Figure~\ref{fig:fmeca_training_plots} shows the training plots with the FMECA utility and Figure~\ref{fig:threshold_training_plots} for the Threshold utility. In summary, MO-DCMAC outperforms the baseline policies in almost all the experiments, with the only exception being the Larger Quay Wall Asset with the Threshold utility, which we will explain later in this section. Moreover, the YBA policies perform significantly worse than MO-DCMAC and the other baseline policies due to them being too passive and not immediately reacting to failing components. When we compare the inspection-based policies, YBI-CBA and CBI-CBA, to each other, we notice that CBI-CBA consistently performs better than YBI-CBA, with the results only comparable at the Simple Homogenous Asset.

For the Simple Homogenous Asset environment, our results indicate that MO-DCMAC achieves the best utility score with both utility functions. When we compare the results of both objectives, cost, and probability of collapse, we notice that MO-DCMAC achieves a significantly lower cost (FMECA: 1.471 ± 0.323, Threshold: 1.498 ± 0.325) than the inspection-based policies YBI-CBA (FMECA: 1.636 ± 0.319, Threshold: 1.651 ± 0.304) and CBI-CBA (FMECA: 1.623 ± 0.338, Threshold: 1.615 ± 0.339), with approximately a 10\% difference for both utility functions. In turn, the probability of collapse is slightly higher with MO-DCMAC (FMECA: 0.059 ± 0.052, Threshold: 0.058 ± 0.051) compared to CBI-CBA (FMECA: 0.056 ± 0.052, Threshold: 0.054 ± 0.052) and YBI-CBA (FMECA: 0.058 ± 0.061, Threshold: 0.053 ± 0.057). However, this difference is almost insignificant. This indicates that MO-DCMAC can learn how to balance cost versus probability of collapse by not performing costly maintenance if it only decreases the probability of collapse by a small amount. Another interesting result with the Simple Homogenous Asset environment is that the difference in results between MO-DCMAC with the Threshold utility and the FMECA utility is insignificant. 

In Table~\ref{table:all_results}, we see that the scores of both objectives, cost and probability of collapse, are almost identical, with MO-DCMAC with the FMECA utility getting 1.471 ± 0.323 for cost, and with the Threshold getting 1.498 ± 0.325, while for the probability of collapse, MO-DCMAC with the FMECA utility is 0.059 ± 0.052 and with the Threshold 0.058 ± 0.051. The most likely cause is that, due to this environment being less complex, with no dependencies between the components, fewer components, and all the components of the same type, there is less flexibility in finding an optimized maintenance plan for this kind of asset. Therefore, it is likely that other utility functions, if correctly formulated, should not lead to different maintenance strategies. This, unfortunately, invalidates one benefit of MO-DCMAC for this environment over other methods, namely, adjusting the utility function should result in different maintenance plans.

With the second environment, the Quay Wall Asset, we see in Table~\ref{table:all_results} that MO-DCMAC again achieves the best utility score with both utilities. Moreover, the difference between MO-DCMAC  (FMECA: 5.072 ± 3.158, Threshold: 1.133 ± 1.186) and CBI-CBA  (FMECA: 5.566 ± 3.544, Threshold: 1.505 ± 0.492), the second-best policy, is larger in this environment than in the Simple Homogenous Asset environment. For instance, the difference in utility score with the Quay Wall Asset environment is 9.34\% with the FMECA utility and 33.42\% with the Threshold utility, whereas with the Simple Homogenous Asset environment, this difference is only 1.64\% with the FMECA utility and 6.94\% with the Threshold utility. This difference shows that MO-DCMAC can outperform other policies by a larger margin when the maintenance problem becomes more complex because it can likely learn the dependencies between components. 

When we focus on the results of each utility, we see that MO-DCMAC achieves the best utility score and cost for both utilities. We see that the difference in cost between MO-DCMAC (FMECA: 1.028 ± 0.156, Threshold: 0.931 ± 0.15) and CBI-CBA (FMECA: 1.49 ± 0.227, Threshold: 1.479 ± 0.236) is substantial. Moreover, with the FMECA utility, the difference between the probability of collapse of MO-DCMAC (0.015 ± 0.02) and CBI-CBA (0.01 ± 0.021) is only 0.005, indicating that MO-DCMAC learns that lowering the probability of collapse after a specific value is less valuable for a better FMECA score. 

The results of the Threshold utility function for the Quay Wall Asset environment show an even lower cost with MO-DCMAC (0.931 ± 0.15) compared to CBI-CBA (1.479 ± 0.236); however, the probability of collapse of MO-DCMAC (0.028 ± 0.033) is significantly higher than of CBI-CBA (0.01 ± 0.018). This difference is because MO-DCMAC learned it should not do certain maintenance actions if it decreases the probability of collapse beyond the threshold. Nevertheless, we see in Table~\ref{table:all_results} that the probability of collapse is, on average, always under the threshold of 0.1. This is true for all the environments, which indicates that during the training process, MO-DCMAC wants to learn a policy that can also handle unforeseen failures that can suddenly increase the probability of collapse. Still, it does not consistently achieve a probability of collapse below the threshold, with the utility score always being higher than the cost. This effect is especially noticeable in the Quay Wall Asset environment with MO-DCMAC (1.133 ± 1.186) because the standard deviation is almost equal to the mean utility score, while this is not the case with the other policies. Therefore, with MO-DCMAC, the probability of being above the threshold is higher than with other policies. A solution might be to increase the penalty if the probability of collapse is above the threshold.

In the last environment, the Larger Quay Wall Asset, we see in Table~\ref{table:all_results} that only the FMECA utility MO-DCMAC (7.293 ± 4.74) can outperform the other policies, such as CBI-CBA (10.806 ± 7.183). The difference in utility score with FMECA between MO-DCMAC and CBI-CBA, the second best-performing policy with FMECA, is larger than with the other environments, with a difference of 46.165\%. When we compare the results of the objectives, we notice that MO-DCMAC can achieve the lowest cost (1.44 ± 0.284) and the second lowest probability of collapse (0.022 ± 0.028).

In contrast, the results of the Larger Quay Wall Asset environment with the Threshold utility show that MO-DCMAC had a significantly worse utility score (3.264 ± 4.114) than CBI-CBA (2.214 ± 2.168). This is the only instance where MO-DCMAC's utility score is worse than that of other policies. The most probable reason is that MO-DCMAC cannot converge to a stable policy with the Threshold utility function for the Larger Quay Wall environment. We can see in training plots in Figure~\ref{fig:threshold_training_plots} that the standard deviation is still noticeable. When we compare this with the training plots of the Larger Quay Wall Asset environment with the FMECA utility in Figure~\ref{fig:fmeca_training_plots}, we see that MO-DCMAC converges to a stable policy. If we compare the training plots in Figure~\ref{fig:fmeca_training_plots} and Figure~\ref{fig:threshold_training_plots} for the other environment, we see that with the Threshold utility, it takes significantly more training steps before MO-DCMAC converges to a stable policy, compared to FMECA. This difference can be attributed to the fact that with the FMECA utility, MO-DCMAC receives more signals about when it performs well because there is always a benefit to lowering the probability of collapse and cost, while with the Threshold utility, it only does so with cost.

Another probable cause for the lower performance with the Larger Quay Wall environment could be the neural network architecture used. The current network architecture is a feedforward network, with one shared input layer for all components and an output head for each component. Due to the increased number of components in this environment, the shared layer struggles with learning each interaction. Increasing the size of the shared input layer might alleviate this, but we do not believe this is the best solution. DDMAC \cite{ANDRIOTIS_paper_2} did not use a shared layer and used separate input layers for each component. This network setup is feasible for the Simple Homogenous Asset environment. Still, it would likely not perform well for the other environments because each maintenance action is taken independently of the states of the other components, and the probability of collapse depends on other components in the same sub-dependency group. 

We believe the best solution would be to change the shared input layer to a Graph Neural Network layer, such as GCN \cite{GCN}. In it, the nodes would be the components, and we would connect nodes in the same sub-dependency group. Using a Graph Neural Network layer would allow each component's action to consider other components in the same sub-dependency group but not the other components.

\begin{figure*}[ht]
    \centering
\includegraphics[width=\textwidth]{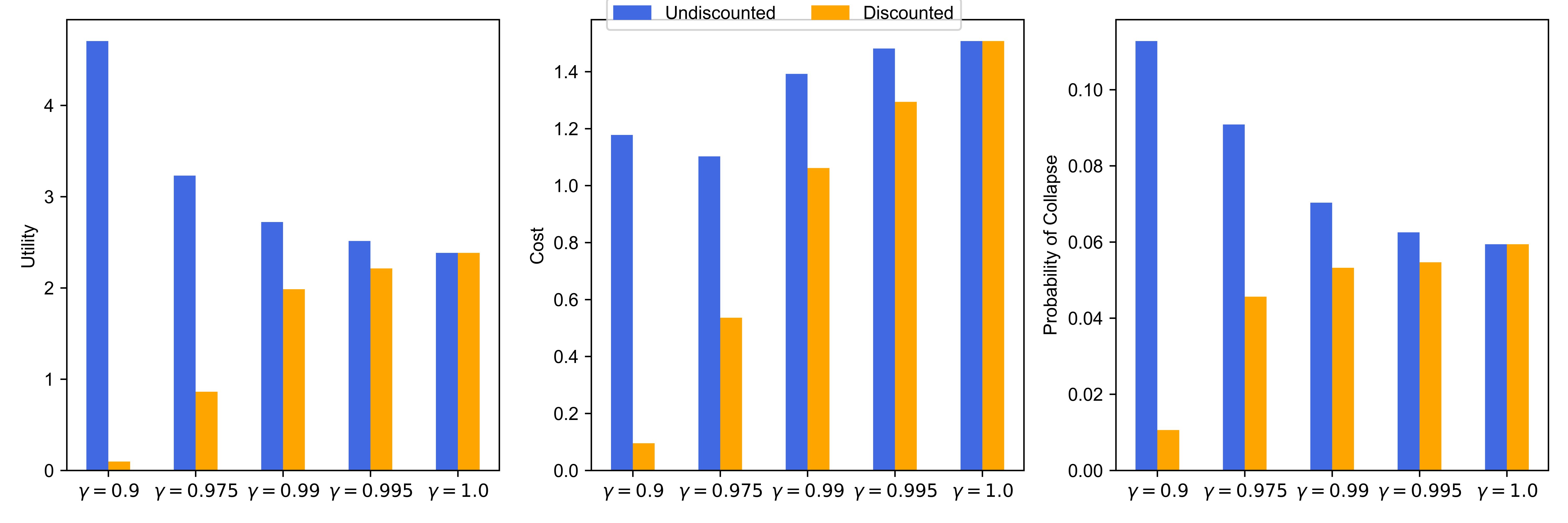}
    \caption{The figure shows the effect of using a different $\gamma$ with the Threshold utility function with the Simple Homogenous Asset environment. The blue bar is the non-discounted value, and the orange bar is the discounted value, which is the value if we apply the corresponding $\gamma$. For each of the plots, a lower score is always strictly better.}
    \label{fig:threshold_diff_gamma_exp}
\end{figure*}
In Section \ref{sec:uti_threshold}, we stated how we wanted to show the effect on the Threshold utility with different discount factors. We argued that a low discount factor might result in a sub-optimal policy due to the difference between the discounted and non-discounted returns, which can have an effect on the utility score. For these experiments, we used the Simple Homogenous Asset environment with the following discount factors: $\gamma=0.9$, $\gamma=0.975$, $\gamma=0.99$, $\gamma=0.995$, and $\gamma=1$

The results in Figure~\ref{fig:threshold_diff_gamma_exp} show how, with the Threshold utility function, a higher discount factor positively correlates with the utility score. Moreover, the right plot in Figure~\ref{fig:threshold_diff_gamma_exp} signifies that with lower values of $\gamma$, the learned policy will result in a higher average probability of collapse. The primary reason for this is that due to the discounting, MO-DCMAC believes it has reached the threshold during training with greater ease and should not put more effort into lowering the probability of collapse. We see this in the middle plot of Figure~\ref{fig:threshold_diff_gamma_exp}, where we see that increasing $\gamma$ results in higher cost, with $\gamma=0.9$ being the exception, likely due to this value of $\gamma$ not being optimal for cost.

These experiments show the importance of selecting a correct $\gamma$ based on the utility function when using MO-DCMAC. Especially if the utility function contains a strict threshold, meaning either above or below a specific objective value, the utility score is changed and not continuously. Another option would be to use a constrained MDP \cite{gu2023review}, where the threshold objective is a constraint that should be satisfied. This approach has other benefits when combined with either MO-DCMAC or MOCAC, namely the inability of these methods to scale to larger sets of objectives due to the output of the critic scaling exponentially to the number of objectives. Therefore, if MO-DCMAC is combined with constrained MDP, we could model the probability of collapse as a constraint instead of an objective. This would allow other objectives in its place, such as the availability of the asset. Moreover, other research shows the application of constrained MDP in infrastructural maintenance optimization \cite{ANDRIOTIS_paper_2}. 

\section{Conclusion} \label{sec:conclusion}
In this paper, we introduced MO-DCMAC, a utility-based MORL approach that can learn to plan maintenance for a multi-component asset. Our results show how MO-DCMAC can outperform commonly used heuristics for maintenance optimization in multiple environments. 

We showed that a utility-based MORL approach for infrastructural maintenance optimization shows promise. We show that MO-DCMAC can learn a policy with a utility based on the FMECA methodology, which is currently used by asset managers to asses maintenance plans. This will allow the learned policy to match the requirements set by those asset managers closely. Moreover, MO-DCMAC can learn different policies depending on the utility used. This is essential because asset managers might need to adjust the utility function because of stronger constraints on the budget or different safety requirements. 

MO-DCMAC was able to show these results in maintenance environments with real-world assets based on deterioration models and interaction between components. However, for future research, we might consider using surrogate models as a basis for a maintenance optimization environment \cite{surrogate}. These surrogate models can more accurately model the infrastructural assets; therefore, the resulting maintenance policy learned by MO-DCMAC will also be more accurate when implemented in the real world. Besides maintenance optimization, we believe that MO-DCMAC can also be used for other multi-objective problems requiring a factorized action space with a known utility function. 

In this paper, we only considered the known utility function scenario described by Hayes et al. \cite{Hayes2022}. However, adapting MO-DCMAC to the dynamic utility function scenario has major benefits because asset managers or policymakers might adjust the thresholds in the Threshold utility or the weighting in the FMECA utility due to new safety requirements or budget constraints. Having MO-DCMAC be able to learn an optimal policy for these dynamic priorities might be more desired than our current approach. Nevertheless, it is essential that the objectives can still be weighted using a non-linear utility under the ESR criterion in this scenario.

MO-DCMAC's major limitation is its inability to scale to more objectives. The number of objectives is due to how the critic's output scales exponentially with the number of objectives. This limited us to using only two objectives in this paper; however, if we look at how maintenance is planned, we see that in most instances, significantly more objectives are considered, such as the availability of the structure. Therefore, the ability to scale up the number of objectives would increase the likelihood that these methods, such as MO-DCMAC, can be used in the real world drastically. We must investigate other approaches to accommodate more objectives \cite{hayes2022multiobjective}.

\backmatter

\begin{appendices}
\section{Environments}
In this appendix, we state the details of each environment we used in our experiments. In each subsection, we state the components, their degradation matrix, the cost of each action, and the probability of collapse when a component or a group of components fails. In this first section, we note the implementation details shared between each environment.

In each environment, the state space of the components consists of the health state, which can be in one of 5 states, and the degradation rate $\tau$, which starts at 0 and with $\tau_{\max}=50$. If a component $\comp_{i}$ is in the last state, it is considered failing.

Each environment has the same available actions for both the global actions and the component actions. The global actions are as follows: \textit{do nothing} and \textit{inspect}. The component actions are: \emph{do nothing}, \emph{repair}, and \emph{replace}. If a \emph{repair} action is taken, the health state is improved with one step, and the current degradation rate does not change. With a \emph{replace} action, the component is set to the best health state, and the degradation rate is reset at the next time step $\tau^{\comp_{i}}_{t+1}=0$. Lastly, if a \textit{do nothing} action is done for the component, the component health state can stay the same or transition to a worse.

This transition probability is described in a degradation matrix $D \in \mathbb{R}^{K\times\tau_{\max}\times5\times5}$, where $K$ is the different number of component groups and 
$\tau_{\max}$ the maximum degradation rate. For the degradation matrix of each component group $k \in K$, we only state the initial degradation matrix when $\tau=0$ and the degradation matrix when $\tau=\tau_{\max}$. The other degradation matrices are interpolated through the following formula:

\begin{align}
    D_{k, \tau=i} =& D_{k,\tau=0} + \frac{i}{\tau_{\max}- 1}\nonumber \\
    &\cdot \left(D_{k,\tau=\tau_{\max}} - D_{k, \tau=0} \right)\nonumber \\
&\forall k \in K \wedge \forall i \in \{0, ..., \tau_{\max}-1\}\label{eq:transition_interpolation}
\end{align}

Each environment also has the same observation space, whereby both the global and component actions influence the observation received. If both the global action and the component action for $\comp_i$ are \textit{do nothing}, the observation is as follows for component $\comp_i$:
\begin{equation}
O^{\textit{do nothing}}=\begin{bmatrix}
1 & 0 & 0 & 0 & 0 \\
1 & 0 & 0 & 0 & 0 \\
0 & 0 & 0 & 0 & 1 \\
0 & 0 & 0 & 0 & 1 \\
0 & 0 & 0 & 0 & 1 \\
\end{bmatrix}.
\end{equation}
However, if either the global action is \textit{inspect} or the component action for $\comp_{i}$ is \textit{repair} or \textit{replace}, the observation will fully reveal the state and is, therefore:
\begin{equation}
    O^{\textit{reveal}}= I_{5}.
\end{equation}

Lastly, at the start of the episode, the belief state of each component will be set to:
\begin{equation} \label{eq:initial_belief}
    \mathbf{b}_{0} = \left \{0.2,0.2,0.2, 0.2, 0.2 \right \}
\end{equation}

\subsection{Simple Homogeneous Asset} \label{appendix:env:simple}
The Simple Homogeneous Asset environment consists of $N=8$ components. 
Each component $\comp_{i}$ can either be in one of 5 states, with each state being progressively worse than the previous state. When the component is in the last state, the component is failing. At each time step, a component $\comp_{i}$ can transition to a worse state. This transition probability is stated in the following degradation matrices, with the initial degradation matrix being:
\begin{equation}
D_{1,\tau=0}=\begin{bmatrix}
0.97 & 0.015 & 0.01 & 0.004 & 0.001 \\
0 & 0.98 & 0.012 & 0.005 & 0.003 \\
0 & 0 & 0.981 & 0.018 & 0.01 \\
0 & 0 & 0 & 0.985 & 0.015 \\
0 & 0 & 0 & 0 & 1 \\
\end{bmatrix},
\end{equation}
and the final degradation matrix being:
\begin{equation}
D_{1,\tau=\tau_{\max}}=\begin{bmatrix}
0.9 & 0.05 & 0.03 & 0.015 & 0.005 \\
0 & 0.91 & 0.06 & 0.02 & 0.01 \\
0 & 0 & 0.92 & 0.06 & 0.02 \\
0 & 0 & 0 & 0.93 & 0.07 \\
0 & 0 & 0 & 0 & 1 \\
\end{bmatrix},
\end{equation}
The degradation matrices for $0 < \tau < \tau_{\max}$ are calculated by Equation~\ref{eq:transition_interpolation}. At the start of the episode, each component starts at its best health state.

For each components $\comp_{i}$, we have three actions, namely 
\emph{do nothing}, \emph{repair}, and \emph{replace}. Each has the following costs: 0, 0.0125, and 0.03125, respectively. There are also two global actions: \emph{do nothing} and \emph{inspect}, whereby the costs are 0 and 0.02.

Lastly, there is no interaction between components when the probability of collapse is calculated because each component belongs to an individual component group. Therefore, if a single component fails, the contribution to the probability of collapse is 0.05.

\subsection{Quay Wall Asset} \label{appendix:env:normal}
The second environment models a simplified version of a section of a historical quay wall in Amsterdam. In it, the goal is to plan maintenance for the wooden components of a section of a quay wall. The environment consists of 13 components and three different component types. The first group is the wooden poles, which are comprised of nine poles, $\text{poles}=\{ \comp_{1},...\comp_{9}\}$. The degradation matrices for the poles are: 
\begin{align} 
&D_{\text{pole},\tau=0} = \nonumber \\
&\begin{bmatrix}
0.983 & 0.0089 & 0.0055 & 0.0025 & 0.0001 \\
0 & 0.9836 & 0.0084 & 0.0054 & 0.0026 \\
0 & 0 & 0.9862 & 0.0084 & 0.0054 \\
0 & 0 & 0 & 0.9917 & 0.0083 \\
0 & 0 & 0 & 0 & 1 \\
\end{bmatrix}\label{eq:begin_deg_poles}
\end{align}
and,
\begin{align}  
&D_{\text{pole},\tau=\tau_{\max}} = \nonumber \\
&\begin{bmatrix}
0.9713 & 0.0148 & 0.0093 & 0.0045 & 0.0001 \\
0 & 0.9719 & 0.0142 & 0.0093 & 0.0046 \\
0 & 0 & 0.9753 & 0.0153 & 0.0094 \\
0 & 0 & 0 & 0.9858 & 0.0142 \\
0 & 0 & 0 & 0 & 1 \\   
\end{bmatrix}\label{eq:end_deg_poles} 
\end{align} 

The second group are the three wooden kesps, $\text{kesps}=\{\comp_{10}, \comp_{11}, \comp_{12}\}$, which have the following degradation matrices:
\begin{align}
&D_{\text{kesp},\tau=0} = \nonumber \\
&\begin{bmatrix}
0.9748 & 0.013 & 0.0081 & 0.004 & 0.0001 \\
0 & 0.9754 & 0.0124 & 0.0081 & 0.0041 \\
0 & 0 & 0.9793 & 0.0125 & 0.0082 \\
0 & 0 & 0 & 0.9876 & 0.0124 \\
0 & 0 & 0 & 0 & 1 \\   
\end{bmatrix} \label{eq:begin_deg_kesp}  
\end{align}
and
\begin{align}
\label{eq:end_deg_kesp}   
&D_{\text{kesp},\tau=\tau_{\max}} = \nonumber\\
&\begin{bmatrix}
0.9534 & 0.0237 & 0.0153 & 0.0075 & 0.0001 \\
0 & 0.954 & 0.0231 & 0.0152 & 0.0077 \\
0 & 0 & 0.9613 & 0.0233 & 0.0154 \\
0 & 0 & 0 & 0.9767 & 0.0233 \\
0 & 0 & 0 & 0 & 1 \\  
\end{bmatrix}.
\end{align}
The last component, $\comp_{13}$, is the wooden floor, for which the degradation matrices are:
\begin{align}
&D_{\text{floor},\tau=0} = \nonumber \\
&\begin{bmatrix}
0.9848 & 0.008 & 0.0049 & 0.0022 & 0.0001 \\
0 & 0.9854 & 0.0074 & 0.0048 & 0.0024 \\
0 & 0 & 0.9876 & 0.0075 & 0.0049 \\
0 & 0 & 0 & 0.9926 & 0.0074 \\
0 & 0 & 0 & 0 & 1 \\  
\end{bmatrix} \label{eq:begin_deg_floor}
\end{align}
and
\begin{align}
&D_{\text{floor},\tau=\tau_{\max}} = \nonumber \\
&\begin{bmatrix}
0.9748 & 0.013 & 0.0081 & 0.004 & 0.0001 \\
0 & 0.9754 & 0.0124 & 0.0081 & 0.0041 \\
0 & 0 & 0.9793 & 0.0125 & 0.0082 \\
0 & 0 & 0 & 0.9876 & 0.0124 \\
0 & 0 & 0 & 0 & 1 \\   
\end{bmatrix}\label{eq:end_deg_floor} .
\end{align}

The cost of each maintenance action can be found in Table~\ref{tab:env_2_prices}. The cost of the global actions is 0 for \textit{do nothing} and 0.02 for \textit{inspection}.

\begin{table}[h]
    \centering
    \caption{The cost of each maintenance action in the Quay Wall environment.}
    \label{tab:env_2_prices}
    \begin{tabular}{cccc}
    \toprule
        \textbf{Component} & \textbf{Do Nothing} & \textbf{Repair} & \textbf{Replace} \\
        \midrule
        Pole & 0 & 0.011 & 0.044 \\
        Kesp & 0 & 0.003 & 0.013 \\
        Floor & 0 & 0.028 & 0.113 \\
        \botrule
    \end{tabular}
    
\end{table}

In this environment, we do consider interaction with component groups. The first type of component groups are wooden poles that influence each other. We define the following groups of this type:
\begin{align}
        \gr_{0}^{\text{poles}}&=\{\comp_0, \comp_1, \comp_2\}\nonumber  \\
        \gr_{1}^{\text{poles}}&=\{\comp_3, \comp_4, \comp_5\} \nonumber \\
        \gr_{2}^{\text{poles}}&=\{\comp_6, \comp_7, \comp_8\}.
\end{align}
When one or more component fails in these groups, the effect on the probability of collapse is found in Table~\ref{tab:prob_collapse_pole}.
\begin{table}[h]
    \centering
    \caption{The effect on the probability of collapse if one or more poles fail in the same sub-dependency groups.}
    \label{tab:prob_collapse_pole}
    \begin{tabular}{cc}
    \toprule
        \textbf{Components Fail} & \textbf{Probability of Collapse} \\
        \midrule
        0 & 0  \\
        1 & 0.01  \\
        2 & 0.1  \\
        3 & 0.4 \\
        \botrule
    \end{tabular}
\end{table}
The second sub-dependency groups are the kesps that affect each other. These are defined as follows:
\begin{align}  
        \gr_{0}^{\text{kesps}}&=\{\comp_{9}, \comp_{10}\} \nonumber\\
        \gr_{1}^{\text{kesps}}&=\{\comp_{10}, \comp_{11}\}.
\end{align}

The effect of the probability of collapse is defined in Table~\ref{tab:prob_collapse_kesp}.
\begin{table}[h]
    \centering
    \caption{The effect on the probability of collapse if one or more kesps fail in the same sub-dependency groups.}
    \label{tab:prob_collapse_kesp}
    \begin{tabular}{cc}
    \toprule
        \textbf{Components Fail} & \textbf{Probability of Collapse} \\
        \midrule
        0 & 0  \\
        1 & 0.03  \\
        2 & 0.33  \\
        \botrule
    \end{tabular}
\end{table}
Lastly, we consider the wooden floor, which itself belongs to a single sub-dependency group:
\begin{equation}
\gr_{0}^{\text{floor}}=\{\comp_{12}\}.
\end{equation}

With the effect of it defined in Table~\ref{tab:prob_collapse_floor}.
\begin{table}[h]
    \centering
    \caption{The effect on the probability of collapse if the floor fails in the same sub-dependency groups.}
    \label{tab:prob_collapse_floor}
    \begin{tabular}{cc}
    \toprule
        \textbf{Components Fail} & \textbf{Probability of Collapse} \\
        \midrule
        0 & 0  \\
        1 & 0.05  \\
        \botrule
    \end{tabular}
\end{table}
Due to insufficient maintenance over the last years, we assume the components are in states 3 and 4 just before the failed state (state 5). The starting state of the components is as follows:
\begin{equation}
    s_0 = \left \{4, 4, 3, 4, 3, 3, 4, 3, 4, 3, 4, 3, 3 \right \}
\end{equation}

\subsection{Larger Quay Wall Asset} \label{appendix:env:difficult}
The last environment, Larger Quay Wall, is similar to the previous environment, Quay Wall, with the major difference being that the number of components is doubled to $N=26$. Therefore, we have the same component groups of wooden poles, kesps, and floors. The wooden poles are $\{\comp_1, ...,\comp_{18}\}$, with the degradation matrices found in Equations~\ref{eq:begin_deg_poles} and \ref{eq:end_deg_poles}, the wooden kesps are $\{\comp_{19}, ...,\comp_{24}\}$ with the degradation matrices found in Equations~\ref{eq:begin_deg_kesp} and \ref{eq:end_deg_kesp}, and the wooden floors are $\{\comp_{25}, \comp_{26}\}$ with the degradation matrices found in Equations~\ref{eq:begin_deg_kesp} and \ref{eq:end_deg_kesp}. The cost of each maintenance action is found in Table~\ref{tab:env_3_prices}.

\begin{table}[h]
    \centering
    \caption{The cost of each maintenance action in the Larger Quay Wall environment.}
    \label{tab:env_3_prices}
    \begin{tabular}{cccc}
    \toprule
        \textbf{Component} & \textbf{Do Nothing} & \textbf{Repair} & \textbf{Replace} \\
        \midrule
        Pole & 0 & 0.0055 & 0.022 \\
        Kesp & 0 & 0.0015 & 0.0065 \\
        Floor & 0 & 0.014 & 0.0565 \\
        \botrule
    \end{tabular}
\end{table}

The sub-dependency groups are as follows for the wooden poles:
\begin{align}
        \gr_{0}^{\text{poles}}&=\{\comp_0, \comp_1, \comp_2\}\nonumber \\
        \gr_{1}^{\text{poles}}&=\{\comp_3, \comp_4, \comp_5\}\nonumber \\
        \gr_{2}^{\text{poles}}&=\{\comp_6, \comp_7, \comp_8\}\nonumber \\
        \gr_{3}^{\text{poles}}&=\{\comp_9, \comp_{10}, \comp_{11}\}\nonumber \\
        \gr_{4}^{\text{poles}}&=\{\comp_{12}, \comp_{13}, \comp_{14}\} \nonumber \\
        \gr_{5}^{\text{poles}}&=\{\comp_{15}, \comp_{16}, \comp_{17}\}.
\end{align}
The sub-dependency groups for the wooden kesps:
\begin{align}  
        \gr_{0}^{\text{kesps}}&=\{\comp_{18}, \comp_{19}\}\nonumber \\
        \gr_{1}^{\text{kesps}}&=\{\comp_{19}, \comp_{20}\}\nonumber \\
        \gr_{2}^{\text{kesps}}&=\{\comp_{20}, \comp_{21}\}\nonumber \\
        \gr_{3}^{\text{kesps}}&=\{\comp_{21}, \comp_{22}\}\nonumber \\
        \gr_{4}^{\text{kesps}}&=\{\comp_{22}, \comp_{23}\}.
\end{align}
Lastly, the wooden floors are independent of each other; therefore, the sub-dependency groups are:
\begin{align}
        \gr_{0}^{\text{floor}}&=\{\comp_{24}\}\nonumber\\
        \gr_{1}^{\text{floor}}&=\{\comp_{25}\}.
\end{align}
The effect of the probability of collapse is the same as the Quay Wall environment and can be found in Tables \ref{tab:prob_collapse_pole}, \ref{tab:prob_collapse_kesp} and \ref{tab:prob_collapse_floor}.

Similar to in the Quay Wall environment, we set the initial start state of the components to a worse state (State 3 or 4):
\begin{align}
    s_{o} = \{&4, 4, 3, 4, 3, 3, 4, 3, 4, 4, 4, 3, 4, 3, \nonumber \\ &3, 4, 3, 4, 3, 4, 3, 3, 4, 3, 3, 4 \}
\end{align}
\end{appendices}
\bmhead{Data availability} The code and data used in the current paper are available from the corresponding author on reasonable request.
\section*{Declarations}
\bmhead{Conflict of interest}
The authors declare that they have no known competing financial interests or personal relationships that could have appeared to influence the work reported in this paper.

\bibliography{bib}
\end{document}